\documentclass[conference]{IEEEtran}
\IEEEoverridecommandlockouts

\usepackage{cite}
\usepackage{amsmath,amssymb,amsfonts}
\usepackage{graphicx}
\usepackage{textcomp}
\usepackage{xcolor}
\usepackage{algorithm}
\usepackage{algpseudocode}
\usepackage{booktabs}
\usepackage{url}
\usepackage{hyperref}
\usepackage{caption}
\usepackage{subcaption}
\usepackage{wrapfig}
\usepackage{comment}
\usepackage{tikz}
\usepackage{pgffor}
\usepackage{tabularx}
\usepackage{multirow}
\usepackage{pifont}
\usepackage{mathtools}
\usepackage{amsthm}
\usepackage{tcolorbox}

\usepackage[textsize=tiny]{todonotes}

\usepackage{graphicx} 
\usepackage{tabularx} 

\newcommand{\Huzaifa}[1]{\textbf{\color{blue}
[Huzaifa: #1]}}

\newcommand{\vC}{{\mathbf{C}}}

\newcommand{\vE}{{\mathbf{E}}}

\newcommand{\vH}{{\mathbf{H}}}

\newcommand{\vT}{{\mathbf{T}}}

\newcommand{\vW}{{\mathbf{W}}}

\newcommand{\vb}{{\mathbf{b}}}

\newcommand{\vn}{{\mathbf{n}}}

\newcommand{\vp}{{\mathbf{p}}}

\newcommand{\vx}{{\mathbf{x}}}
\newcommand{\vy}{{\mathbf{y}}}

\newcommand{\cL}{{\mathcal{L}}}

\begin{document}

\title{ PEEL the Layers and Find Yourself:\\ Revisiting Inference-time Data Leakage for \\ Residual Neural Networks
}

\author{
\IEEEauthorblockN{Huzaifa Arif\textsuperscript{1}, Keerthiram Murugesan\textsuperscript{2}, Payel Das\textsuperscript{2}, Alex Gittens\textsuperscript{1}, Pin-Yu Chen\textsuperscript{2}}\\
\IEEEauthorblockA{\textsuperscript{1}Rensselaer Polytechnic Institute, Troy, NY, United States\\
\textsuperscript{2}IBM Research, Yorktown Heights, NY, United States \\
Email: arifh@rpi.edu, Keerthiram.Murugesan@ibm.com, daspa@us.ibm.com, gittea@rpi.edu, pin-yu.chen@ibm.com}
}

\maketitle

\begin{abstract}
 This paper explores inference-time data leakage risks of deep neural networks (NNs), where a curious and honest model service provider is interested in retrieving users' private data inputs solely based on the model inference results. Particularly, we revisit residual NNs due to their popularity in computer vision and our hypothesis that residual blocks are a primary cause of data leakage owing to the use of skip connections. By formulating inference-time data leakage as a constrained optimization problem, we propose a novel backward feature inversion method, \textbf{PEEL}, which can effectively recover block-wise input features from the intermediate output of residual NNs. The surprising results in high-quality input data recovery can be explained by the intuition that the output from these residual blocks can be considered as a noisy version of the input and thus the output retains sufficient information for input recovery. We demonstrate the effectiveness of our layer-by-layer feature inversion method on facial image datasets and pre-trained classifiers. Our results show that PEEL outperforms the state-of-the-art recovery methods by an order of magnitude when evaluated by mean squared error (MSE). The code is available at \href{https://github.com/Huzaifa-Arif/PEEL}{https://github.com/Huzaifa-Arif/PEEL}
\end{abstract}


\section{Introduction}
\label{sec:intro}

\textit{What can a curious but honest model service provider know about your data with just one single forward inference?}

With the prevalence and advances of ``vision foundation models'', users can use off-the-shelf pre-trained models as a service to run model inference on their private data without any model tuning and training. For example, a user can upload a private facial image to an online gender classifier to obtain gender predictions. While most model inference services have explicitly stated that the user-provided data will not be stored nor accessed by the service provider, the narrative of ``We won't know anything about your data'' may not prevent a curious but honest model service provider from attempting to reconstruct the input data from the model inference results returned to the user. 


Here, the end user can be a person consuming the technology, or even an AI-empowered agent that aims to call an API to solve some user-specified tasks. In other words, based on the model inference results, such as the class prediction likelihood of a given image, the service provider can attempt to leverage this knowledge to reconstruct the supposedly private user data without altering the inferences. However, in this work, our focus is on reconstructing inputs from the final output layer of residual blocks within ResNet architectures, rather than directly from final model confidences or logits. This scenario is particularly relevant in contexts like split learning, where intermediate representations are exposed during collaborative training processes. Additionally, storing the last residual layer's embeddings is a common practice for model owners, as these embeddings are often utilized for confidence calibration through temperature scaling. Users may adjust temperature values to influence class prediction confidences, making the reconstruction from residual layer outputs a significant privacy concern regardless of such hyperparameter adjustments. (Section \ref{sec:HbC} discusses this scenario in more detail)

\begin{figure}[t] 
    \centering
    \includegraphics[width=0.99\linewidth]{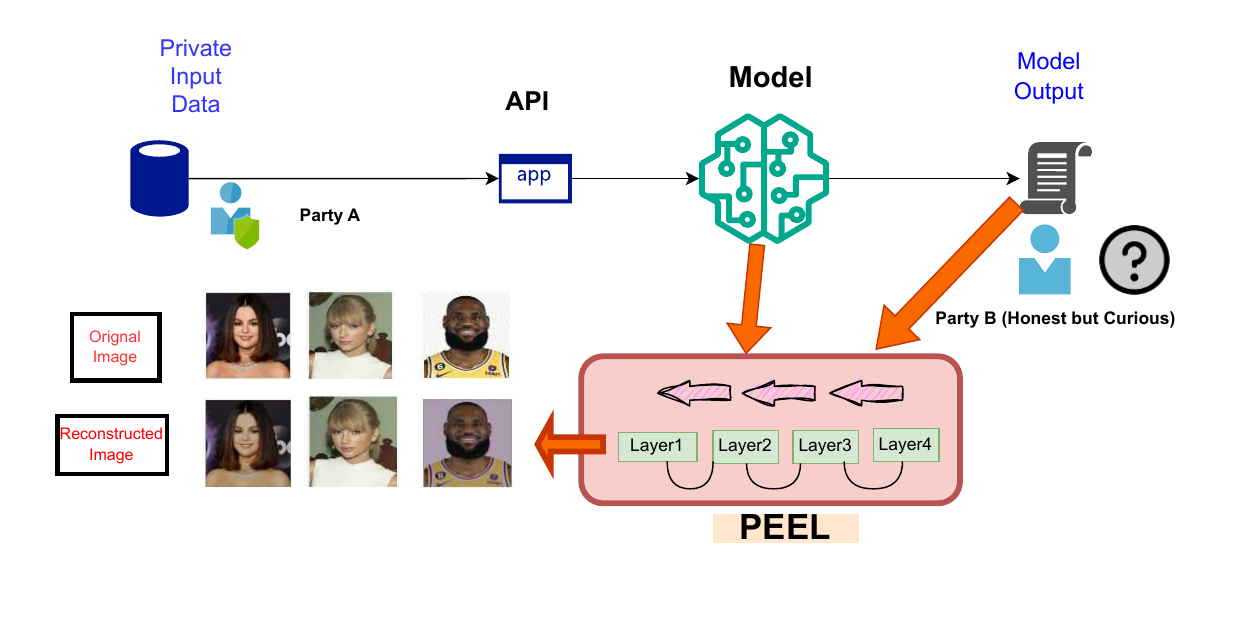} 
        \vspace{-6mm}
        \caption{\textbf{Data leakage problem setup:} Party A (e.g., an enterprise user) accesses an API to make predictions on its private data. Party B (e.g., a model service provider) has access to the model predictions and model weights. It then uses PEEL to reconstruct the private data of Party A. PEEL is an optimization-based feature inversion method, which features block-wise input recovery from the intermediate output of residual neural networks. 
    }
    \label{teaser}
    \vspace{-4mm}
\end{figure}

In this paper, we cast this problem setup as  ``inference-time data leakage'' to explore the risks of data privacy leakage solely based on model inference results. We note that our problem setup differs from standard model inversion attacks at inference time, which aim to reconstruct a data sample used to \textit{train} the model \cite{GMI,VMI,rethink,KEDMI}. It is also distinct from the problem setup of training-time data leakage that studies how to reconstruct training data during model training or fine-tuning \cite{gradient1,gradient2,gradient3,gradient4,gradient5}. As illustrated in Figure \ref{teaser}, we consider the scenario similar to model inference with online APIs, where private data are not used to train the model, and the model weights are fixed when running inference on the private data.
Specifically, this paper studies inference-time data leakage for residual neural networks (ResNets) in image classification. The reasons are that (i) ResNets are relatively mature and widely used backbones for many computer vision tasks. Recent studies also show that ResNets are more competent backbones than vision transformers under the same training conditions \cite{wightman2021resnet,goldblum2023battle}. (ii) ResNets feature residual structures that utilize skip connections to add the input of an intermediate block to its output. We hypothesize that the nature of residual connections facilitates inference-time data leakage, because the block output can be viewed as a noisy version of the block input, when compared to feedforward neural networks without skip connections.

Given a model inference output that is accessible to the model service provider, e.g., the logits of a data sample provided by a ResNet classifier, we propose a novel layer-wise feature inversion method named PEEL, which subsequently reconstructs the intermediate block input from its output layer-by-layer in a backward order to uncover the private data (the input to the first block of the model). It is worth noting that unlike existing methods that require training a generative model on similar (in-distribution) data samples for data reconstruction \cite{GMI,KEDMI,VMI,rethink},
our approach does not require any information other than the model inference results and the model details that are known to the curious but honest service provider. Additionally, such generative approaches are appropriate for recovering approximate versions of the \emph{training} data only, while PEEL facilitates the recovery of arbitrary inference-time data.

We summarize our \textbf{main contributions} as follows.

\begin{figure}[t]
    \centering
    \begin{subfigure}{0.32\textwidth}
        \centering
        \includegraphics[width=\linewidth]{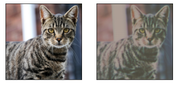}
        \caption{\textbf{Embedding inversion} on randomly initialized ResNet-18 architecture from the embedding of shallow layer (\textbf{Layer 1} in Figure:\ref{PEEL_untrained}).}
        \label{shallow_embedding_inversion}
    \end{subfigure}
    \hfill 
    \begin{subfigure}{0.32\textwidth}
        \centering
        \includegraphics[width=\linewidth]{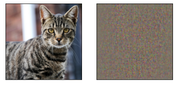} 
        \caption{\textbf{Embedding inversion} on randomly initialized ResNet-18 architecture from the embedding of  deeper layer (\textbf{Layer 4} in Figure:\ref{PEEL_untrained}).}
        \label{deep_embedding_inversion} 
    \end{subfigure}
  \hfill 
      \begin{subfigure}{0.32\textwidth}
        \centering
        \includegraphics[width=\linewidth]{Styles/pictures/Shallow_embedding_inversion.png} 
        \caption{\textbf{PEEL} on randomly initialized ResNet-18 architecture from the embedding of  deeper layer (\textbf{Layer 4} in Figure:\ref{PEEL_untrained}).}
        \label{deep_PEEL_inversion} 
    \end{subfigure}
    \caption{ Comparison of the \textbf{embedding inversion method} \cite{2015_CVPR} and \textbf{PEEL} (ours).
    Deeper layers are hard to invert in ResNets for \cite{2015_CVPR}, whereas \textbf{PEEL} shows good reconstruction performance.
    }
    \label{fig:example}
    \vspace{-4mm}
\end{figure}

\begin{enumerate}
    \item Inspired by the model-agnostic embedding inversion approach \cite{2015_CVPR}, we reformulate the embedding inversion problem into sequential feature reconstruction for residual blocks. We propose a new method called PEEL for input data reconstruction via block-wise backward feature inversion from the model inference results. A visual comparisons is presented in Figure \ref{fig:example}. 
    \item Our results show that using either pretrained weights or randomly initialized weights for model inference can cause severe data leakage of individual images in residual blocks, uncovering a risk of inference-time data leakage inherent to ResNet architectures.
    \item Evaluated on facial recognition tasks and Chest-X ray images, when compared to generative approaches that sample approximate training images, the images recovered by PEEL retain important facial features. Additionally, when measured with regard to popular evaluation metrics such as MSE or PSNR, our method reveals an order of magnitude improvement over the generative approaches. 
\end{enumerate}




\section{Related Work}

Early work on inverting deep representations often relied on direct optimization in pixel space. ~\cite{2015_CVPR} introduced a seminal approach for reconstructing images from intermediate CNN features, demonstrating promising results on shallow or early-layer embeddings. ~\cite{Dosovitskiy2016Inverting} extended these ideas to deeper visual features, while ~\cite{Zhmoginov2016Inverting} specifically investigated inverting embeddings in face-recognition pipelines. In practice, such optimization-based methods typically degrade in reconstruction quality for deeper networks.

Since naive optimization often struggles for highly non-linear or deep embeddings, another line of work leverages pretrained generators to map embeddings back into a latent space. ~\cite{Vendrow2021Realistic} proposed a generative model specifically designed to invert embeddings into realistic face images, and ~\cite{otroshi2023face} employed StyleGAN3 to reconstruct facial templates. Likewise, state-of-the-art model inversion attacks~\cite{GMI,KEDMI,VMI,rethink} often rely on GANs or variational techniques to approximate training distributions, thus recovering ``class-consistent'' or ``representative'' images rather than the \emph{exact} inputs. However, these GAN-based methods require in-distribution data or fine-tuned generators; moreover, their synthesized outputs often differ visibly from real images.

In contrast to prior works that focus on embedding inversion at training time or employ pretrained generators, our work examines inference-time data leakage in \emph{residual blocks}~\cite{invert_resnet}. Residual networks (ResNets) are a widely used architecture for high-dimensional tasks owing to their skip connections, which add the block input to a non-linear transformation of that input (see Figure~\ref{resblock_pic}). Although residual blocks are frequently deemed non-invertible---particularly when weights are noncontractive~\cite{invert_resnet}---we show that substantial input information can still be recovered at inference. Natural images commonly lie on a low-dimensional manifold~\cite{rethink}, complicating direct recovery for deep or non-linear models; hence, most existing model inversion attacks~\cite{GMI,KEDMI,VMI,rethink} rely on generative methods to recover only approximate or class-representative data. By contrast, our work revisits the idea of optimization-based embedding inversion~\cite{2015_CVPR} in the setting of residual networks, exposing more severe privacy vulnerabilities without requiring additional in-distribution data. For ease of discussion, we use ``embedding,'' ``feature,'' and ``image representation'' interchangeably.

\section{Preliminaries}
\label{prelims}

\subsection{Residual Block}

\begin{figure}[htbp]
  \begin{center}
  \includegraphics[width=0.5\linewidth]{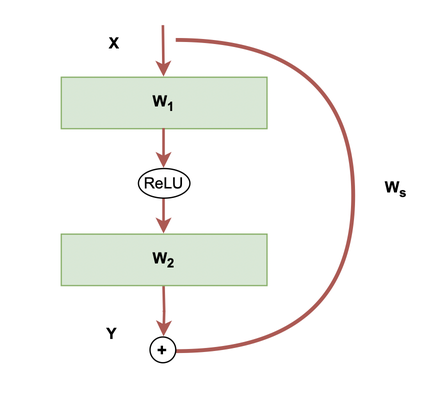}
  \end{center}
  \vspace{-6mm}
  \caption{The structure of the preactivation residual blocks following~\cite{identity_mapping}. Here $\vx$ and $\vy$ are the input and output, respectively, of the residual block. ResNet architectures consist of multiple residual blocks chained in sequence.}
  \label{resblock_pic}
  \vspace{-2mm}
\end{figure}

We consider a residual block  \cite{resnet}; there are many variants of this residual blocks, and we consider a preactivation residual block for this paper \cite{identity_mapping}  with the following formulation in equation (\ref{resblock}):
\begin{align}
\label{resblock}
    \vy = \vW_{s}\vx + \vW_{2} ReLU(\vW_{1}\vx)
\end{align}
Usually, the operation $\vW_{1} \vx$ means a convolution operation. For this work, $\vW_{1},\vW_{2},\vW_{s}$  are 2D convolution operations on multichannel inputs.
$\vW_{s}$ is the downsampling operation to ensure that the dimensions of the skip connections match the output of the convolution operations.
Figure  \ref{resblock_pic} explains the residual block operation.

\subsection{Embedding Inversion \cite{2015_CVPR}}
\label{sec:Embedding}

Recovering input images from a given image representation was first proposed in \cite{2015_CVPR}. The objective in this paper is to formulate the image recovery as a loss function with the authors suggesting a use of Euclidean loss and optimizing over input image space. Consider the equation from \cite{2015_CVPR} that formulates this recovery:
\begin{align}
    \mathbf{x}^{*} = \arg \min_{\mathbf{x} \in \mathbb{R}^{H \times W \times C}} \left\| \Phi(\mathbf{x}) - \Phi_{0}  \right\|_{2}^{2}
\end{align}
where $\Phi_{0}$ is the image representation. The goal of the above equation is to find the image that yields the closest image representation to $\Phi_{0}$. Since this equation is non-convex in general and the goal is to search over a low-dimensional manifold of images, the above equation is modified to give meaningful results as expressed in equation (3) in \cite{2015_CVPR}:

\begin{align}
   \mathbf{x}^{*} &=  \arg \min_{\mathbf{x} \in \mathbb{R}^{H \times W \times C}} \left\| \Phi( \vx) - \Phi_0 \right\|^2 / \left\| \Phi_0 \right\|^2  \nonumber \\
   &\quad + \lambda_{\alpha} R_{\alpha}(\vx) + \lambda_{V\beta} R_{V\beta}(\vx)
   \label{embedding_inv}
\end{align}

The regularizers added here are based on the assumption that the inputs are from a set of natural images. The 
$\alpha\text{-norm}$ regularizer defined here is  $ R_{\alpha}(\vx) = \left\| x \right\|_{\alpha}^{\alpha}$. The other regularizer is for smooth images -- the TV-norm is defined as $R_{V \beta}(\vx) = \sum_{i,j} \left(  \left( (x_{i, j+1} - x_{i, j})^2 + (x_{i+1, j} - x_{i, j})^2 \right) \right)^{\frac{\beta}{2}}$. The hyperparameters choice here is $\alpha = 6$ and $\beta = 2$ as suggested in \cite{2015_CVPR} to recover smooth images.

A noticeable observation from \cite{2015_CVPR} is that it relates the quality of reconstruction to the layer of the image representation $\Phi_{0}$. If the embeddings are taken from a shallow network, this inversion process gives us good reconstruction results while if the embedding is from a deeper layer the quality of reconstruction is very poor. For instance, see the results of shallow and deep layers in Figure~\ref{shallow_embedding_inversion} and \ref{deep_embedding_inversion}. When a deeper embedding is considered from a network, the process of inversion is challenging. 
From a privacy standpoint, it might appear that leaking the embeddings of deep layers is not detrimental to data recovery based on \cite{2015_CVPR}. However,
in the next section, we explain how PEEL overcomes this challenge of deep embedding inversion, as shown in \ref{deep_PEEL_inversion}.

\subsection{Generative Methods for Model Inversion}

Existing approaches in model inversion tasks employ generative approaches to model inversion that leverage distributional assumptions in order to regularize the recovery problem~\cite{GMI,KEDMI,VMI,rethink}. These approaches parameterize the manifold of plausible inputs using a generative model, by assuming that each image can be written as $x= G(z)$ where $z\sim q(z)$ are low-dimensional latent features. 

Given target features, \cite{rethink} \textbf{Rethink-MI} recovers an input image by solving
\begin{equation}
\label{eqn:generative_inversion}
z^\star = \text{argmin}_{z}~L_{\text{prior}}(z) + L_{\text{id}}(z),
\end{equation}
where $L_{\text{prior}}(z) = D(G(z))$ measures the plausibility of the image with latent features $z$, using a GAN trained on an auxiliary public data set, and $L_{\text{id}}(z) = C(G(z))$ is a metric that measures how closely the image matches the features observed from the target network. To more accurately model the class-conditional image distribution, \cite{KEDMI} trained the GAN to discriminate not simply between plausible and implausible images, but also the different target classes of images.

The work~\cite{VMI} \textbf{VMI} advocated a uniform view of  generative model inference attacks as attempting to sample from the class conditional image distribution $p(x|y)$ given knowledge of the target network, which defines the image conditional class distribution $p(y|x)$. Based upon this formal viewpoint,~\cite{VMI} proposed a variational approach to learning $p(x|y)$.
\cite{GMI,KEDMI,VMI} \textbf{GMI, KEDMI, VMI} develop different $L_{\text{prior}}$, but all utilize multiclass logistic loss to measure how closely the image matches the observed features in $L_{\text{id}}(z)$.~\cite{rethink} proposed to instead use an $L_{\text{id}}$ objective that maximizes the logit corresponding to the most probable class label, and showed empirically that this leads to significant gains in attack performance. We compare \textbf{PEEL} with different inversion methods  in Table \ref{baselines}.

\begin{table}[t]
  \caption{PEEL vs. Baselines. The columns illustrate the additional information required by generative methods and embedding inversion (see Section \ref{sec:Embedding})}
  \label{baselines}
  \centering
  \resizebox{\columnwidth}{!}{%
    \begin{tabular}{lcccr}
      \toprule
      Method & Auxiliary Information & GAN  \\
      \midrule
      KEDMI \cite{KEDMI} & $\surd$ & $\surd$  \\
      Embedding Inversion \cite{2015_CVPR} & $\times$ & $\times$  \\
      GMI \cite{GMI} & $\surd$ & $\surd$\\
      VMI \cite{VMI} & $\surd$ & $\surd$  \\
      Rethink-MIA \cite{rethink} & $\surd$ & $\surd$\\
      \textbf{PEEL} (ours) & $\times$ & $\times$  \\
      \bottomrule
    \end{tabular}
  }
  \vspace{-2mm}
\end{table}

\subsection{Adversarial HbC Setting }
\label{sec:HbC}

In our discussion of \textbf{PEEL}, we adopt the $\vH\vb\vC$ (Honest but Curious) framework, a widely-used setting for evaluating privacy risks in machine learning services and distributed learning environments. Here \textit{honest} means the adversary (see Party B in Figure \ref{teaser}) stays faithful to the model predictions returned to the user (Party A in Figure \ref{teaser}) and does not attempt to alter the prediction results.  However, the adversary is also \textit{curious}, meaning it attempts to use these predictions to infer users' input data.

This setting has been rigorously formalized in works such as \cite{split1,split2}. More recent studies, exemplified by \cite{gradient5}, apply the $\vH\vb\vC$ adversarial server model to analyze and defend against gradient-based attacks in federated learning. In this setting, service providers are assumed to follow the prescribed protocols but may still attempt to infer sensitive information from the data they access.

The key assumption in the $\vH\vb\vC$ model is that while participants like service providers are not explicitly malicious, they have both the ability and incentive to extract as much information as possible from the data they encounter. This setting is highly relevant to modern AI applications, especially those that are proprietary and closed-source. For example, services like ChatGPT allow users to opt out of sharing their chat history with OpenAI, and Microsoft's Copilot assures that no chat history is retained when used with commercial data protection. Despite these promises, the $\vH\vb\vC$ model remains pertinent, as it acknowledges the risk that service providers, even if adhering to legal and marketing obligations not to store user inputs, may still exploit cached outputs from neural networks to reconstruct sensitive user data.

In practice, if a service provider has direct access to raw user inputs, they may not need to employ data reconstruction techniques. However, when providers are restricted from storing inputs, either due to regulations or marketing commitments, they may still cache the outputs of models. The $\vH\vb\vC$ concerns arise here because these cached outputs may be exploited to reconstruct user inputs without breaching the stated terms of service. 

Additionally, inversion attacks are a known risk in other contexts such as split learning \cite{hbc1}, where the neural network's architecture is distributed across multiple service providers. One of the main motivations for split learning is enhancing security by ensuring that the user only needs to trust the first provider, who generates the initial embedding of the input. Techniques like Differential Privacy, Secure Multi-Party Computation (SMPC), or Homomorphic Encryption (HE) \cite{hbc2} are often employed to safeguard privacy during training in split learning. However, during inference, residual information can still pose a security risk. Our work underscores this concern, highlighting the vulnerabilities of residual architectures during inference in split learning and emphasizing the need for robust privacy measures. Additional discussion on the $\vH\vb\vC$ setting is included in Appendix \ref{sec:additional HbC}.

Thus, the $\vH\vb\vC$ setting we consider to evaluate  \textbf{PEEL} particularly relevant, as it illustrates how $\vH\vb\vC$ adversaries might leverage \textbf{PEEL} to recover input data when residual architectures are employed. In the following section we delve into the technical details of \textbf{PEEL}.

\section{Methodology of PEEL}
\label{sec:method}

A chain of Residual Blocks (see Figure in Appendix~\ref{resnet18}) forms the backbone of ResNet architectures. Consider an honest-but-curious ($\vH\vb\vC$) adversary employing \textbf{PEEL} to sequentially recover the inputs to these residual blocks. The adversary begins deep in the network and progressively works backwards towards the initial layers. The inversion of the Residual Blocks is achieved through a novel optimization formulation, while the initial convolutional layers are inverted using the method proposed in~\cite{2015_CVPR}.



For \textbf{PEEL}, the attacker requires only knowledge of the weights $\vW = \{\vW_{1},\vW_{2},\vW_{s}\}$ of each residual block; the weights of the initial convolutional layers of the network; and the model output $f(\vx;\vW)$. No information about the training data distribution is known by the attacker or required by \textbf{PEEL}. Table~\ref{baselines} compares the requirements of \textbf{PEEL} to those of prior approaches. We see that, like the approach of~\cite{2015_CVPR} (\textbf{Embedding Inversion}), it does not require auxiliary information and it attempts true inversion to recover inputs at inference-time, as opposed to generating distributionally plausible approximations to the training data in generative methods. Unlike \textbf{Embedding Inversion}~\cite{2015_CVPR}, PEEL works with deep ResNets \ref{fig:example}.

\begin{figure*}[t]
  \begin{subfigure}[b]{0.45\linewidth}
    \centering    \includegraphics[width=\textwidth]{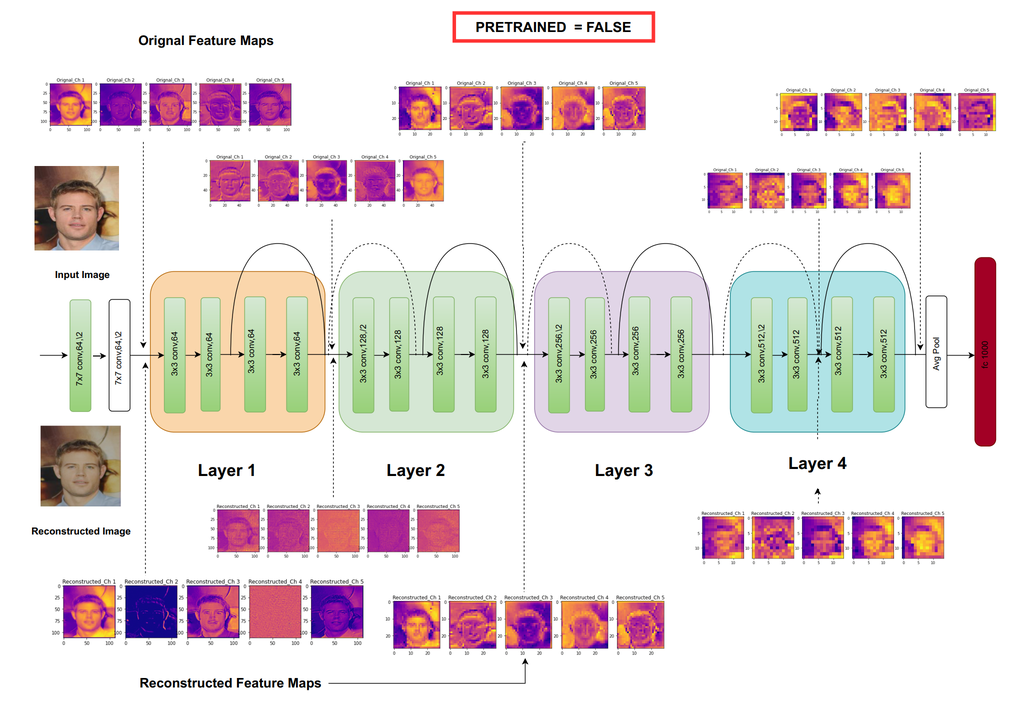}
   \caption{PEEL on Untrained Networks}
    \label{PEEL_untrained}
  \end{subfigure}
  \hfill
  \begin{subfigure}[b]{0.45\linewidth}
    \centering \includegraphics[width=\textwidth]{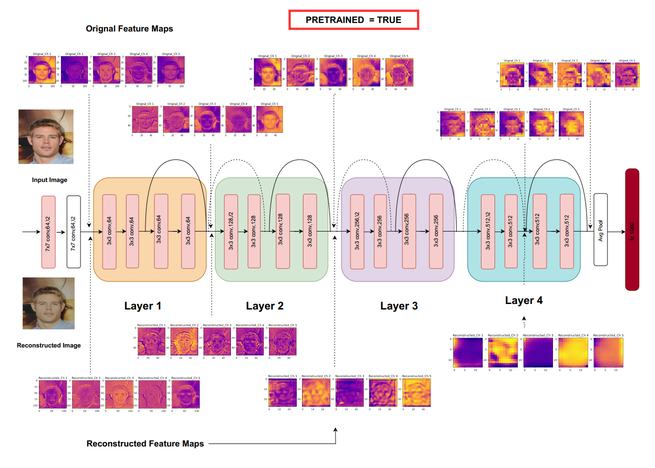}
    \caption{PEEL on Trained Networks}
    \label{PEEL_trained}
  \end{subfigure}
      \vspace{-1mm}
  \caption{PEEL in effect on ResNet 18 architecture. Figure~\ref{PEEL_untrained}/Figure~\ref{PEEL_trained} shows reconstruction on a randomly-initialized/pretrained ResNet18. The intermediate results from each layer visualize the features in a small subset of the channels. The top/bottom half of each figure shows the ground-truth/ reconstructed feature maps using PEEL.}
  \label{fig:face_evolve}
  \vspace{-2mm}
\end{figure*}

\subsection{Feature Inversion for a Residual Block}
The core algorithmic innovation of PEEL is a novel optimization-based approach to recover the inputs to Residual Blocks. Recall the formula describing how a Residual Block maps an input $\vx$ to the corresponding output $\vy$: 
\begin{align*}
    \vy & = \vW_{s}\vx + \vW_{2} ReLU(\vW_{1}\vx) = \vW_{s}\vx + \vW_{2} \vp,
\end{align*}
where we introduced the notation $\vp = ReLU(\vW_{1} \vx)$. Because the attacker only observes the output $\vy$, the vector $\vp$ is unknown. However, the attacker knows that $\vp$ is an entrywise positive image that satisfies
\[
 \vW_1 \vx = ReLU(\vW_{1} \vx) - ReLU(-\vW_{1} \vx) = \vp - \vn,
\]
where $\vn$ is an entrywise positive image. The supports of $\vn$ and $\vp$ are disjoint, so $\vn^T \vp = 0.$ These observations motivate recovering $\vx$ by finding the images $\vx$ and $\vp$ that satisfy these constraints and minimize the squared-error in approximating $\vy$. Thus, \textbf{PEEL} inverts a single residual block by solving the following optimization problem:
 \begin{align}
    \label{resblock_inversion}
    \vx^{*}, \vp^{*},\vn^{*} &= \arg \min_{\vx,\vp,\vn} \| \vy - \vW_{s}\vx - \vW_{2}\vp \|^{2}_{2} \\
    &\text{s.t.} \vW_{1}\vx = \vp - \vn \notag, ~\vn \geq 0 \notag, ~\vp \geq 0 \notag, ~\vn^{T}\vp = 0 \notag
\end{align}

\subsection{PEELing one Residual Block}
\begin{figure}[t]
    \centering
    \begin{subfigure}{0.48\textwidth}
        \centering
        \includegraphics[width=\linewidth]{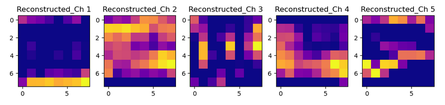}
        \caption{Original Input to Residual Block}
        \label{CIFAR_shallow_orig}
    \end{subfigure}
    \hfill 
    \begin{subfigure}{0.48\textwidth}
        \centering
        \includegraphics[width=\linewidth]{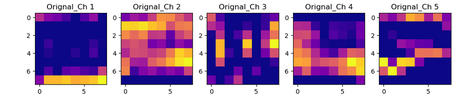} 
        \caption{Reconstructed Input to Residual Block}
        \label{CIFAR_shallow_recons} 
    \end{subfigure}
    \vspace{-1mm}
    \caption{Reconstruction of a 5-channel, 8-by-8 image from the output of a Residual Block using a CIFAR-10 data sample. The residual error in the image recovered from solving~\eqref{resblock_inversion} using PyGRANSO has $\ell_2$-norm 3.64 and 1.14 \% relative error.}
    \vspace{-4mm}
\end{figure}

The optimization problem introduced in Equation ~\eqref{resblock_inversion}  to invert one Residual Block is non-convex and has non-linear constraints. Locally optimal solutions can in principle be obtained using a multitude of non-convex optimization methods in the literature, including \cite{nonconvex1,nonconvex2}. PyGRANSO~\cite{Pygransso}--a popular solver suitable for solving general non-convex optimization problems--is particularly convenient as it integrates into PyTorch and supports autodifferentiation. This solver is efficient and gives small reconstruction errors for low-dimensional inversion problems. Consider, for instance, solving Equation~\eqref{resblock_inversion} to recover a CIFAR-10 image passed through a Residual Block. Results obtained using PyGRANSO are shown in Figures~\ref{CIFAR_shallow_orig} and~\ref{CIFAR_shallow_recons}.

Although PyGRANSO is attractive when solving low-dimensional problems, PyGRANSO becomes computationally infeasible when attempting to recover high-dimensional images. Consider for instance, a residual block of 5 channels and image size of 8 by 8; PyGRANSO takes approximately 30 minutes to solve for input. In comparison, we find that the penalty method takes 40-45 seconds to do similar inversion. Thus, in these scenarios, we employ a penalty method (see Equation ~\eqref{penalty_equation} where  we  solve Equation ~\eqref{resblock_inversion} by finding minimizers of the objective )

\begin{align}    
\label{penalty_equation}
    \cL(\vx,\vp,\vn) = &\ \| \vy - \vW_{s}\vx - \vW_{2}\vp \|^{2}_{2} \nonumber \\
    & + \lambda_{1} \cdot \|\vn^{T}\vp\|_{2}^{2} \nonumber \\
    & + \lambda_{2} \cdot  \|(\vW_{1}\vx - \vp - \vn)\|^{2}_{2}.
\end{align}

A projected gradient descent method is used to handle the nonnegativity constraints on $\vn$ and $\vp$:
\begin{equation}
    \label{penalty_update}
    (\vx^{t+1}, \vp^{t+1}, \vn^{t+1}) = P_{\mathcal{K}} \left( (\vx^{t},\vp^{t},\vn^{t}) -  \eta_t \nabla \mathcal{L}(\vx^{t},\vp^{t},\vn^{t}) \right).
\end{equation}
Here $P_{\mathcal{K}}$ denotes the projection onto the convex cone $\mathcal{K}= \mathbb{R} \times \mathbb{R}_+^{H^\prime \times W^\prime \times C^\prime} \times  \mathbb{R}_+^{H^\prime \times W^\prime \times C^\prime}$, where $H^\prime \times W^\prime \times C^\prime$ is the dimensions of $\vn$ and $\vp$. In practice, the Adam optimizer is used to implement adaptive projected gradient descent. Section~\ref{sec:exp} addresses the choice of hyperparameters $\lambda_1$ and $\lambda_2$.

\subsection{PEELing the entire Residual Network}

\begin{figure*}[t]
    \centering
    \begin{tabular}{>{\centering\arraybackslash}m{0.12\textwidth} >{\centering\arraybackslash}m{0.12\textwidth} >{\centering\arraybackslash}m{0.12\textwidth} >{\centering\arraybackslash}m{0.12\textwidth} >{\centering\arraybackslash}m{0.12\textwidth} >{\centering\arraybackslash}m{0.12\textwidth} >{\centering\arraybackslash}m{0.12\textwidth}}
        \scriptsize\textbf{GROUND TRUTH} & \scriptsize\textbf{KEDMI-LOMMA} & \scriptsize\textbf{Embedding\_Inversion} & \scriptsize\textbf{PEEL (\textbf{U})} & \scriptsize\textbf{PEEL (\textbf{MU})} & \scriptsize\textbf{PEEL (\textbf{P})} & \scriptsize\textbf{PEEL (\textbf{MP})} \\
        \includegraphics[width=0.12\textwidth]{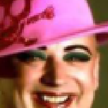} & \includegraphics[width=0.12\textwidth]{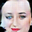} & \includegraphics[width=0.12\textwidth]{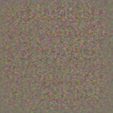} & \includegraphics[width=0.12\textwidth]{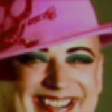} & \includegraphics[width=0.12\textwidth]{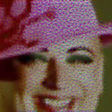} & \includegraphics[width=0.12\textwidth]{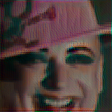} & \includegraphics[width=0.12\textwidth]{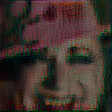} \\
        
        \scriptsize\textbf{GROUND TRUTH} & \scriptsize\textbf{KEDMI-LOMMA} & \scriptsize\textbf{Embedding\_Inversion} & \scriptsize\textbf{PEEL (\textbf{U})} & \scriptsize\textbf{PEEL (\textbf{MU})} & \scriptsize\textbf{PEEL (\textbf{P})} & \scriptsize\textbf{PEEL (\textbf{MP})} \\
        \includegraphics[width=0.12\textwidth]{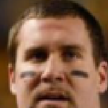} & \includegraphics[width=0.12\textwidth]{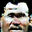} & \includegraphics[width=0.12\textwidth]{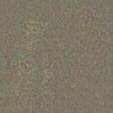} & \includegraphics[width=0.12\textwidth]{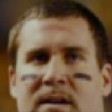} & \includegraphics[width=0.12\textwidth]{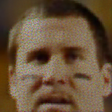} & \includegraphics[width=0.12\textwidth]{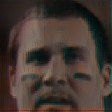} & \includegraphics[width=0.12\textwidth]{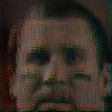} \\
    \end{tabular}
    
    \caption{The first row corresponds to reconstruction results from \textbf{class 1} images. The second row corresponds to reconstruction results from \textbf{class 2} images.}
    \label{fig:reconstruction_comparisons}
    \vspace{-3mm}
\end{figure*}

\begin{algorithm}[t]
\caption{PEEL on ResNet with $N$ Residual Blocks preceded by shallow feature extraction layers.}
{\small
\label{PEEL+embedding_inversion}
\begin{algorithmic}[1]
\Require $\mathbf{y}^{N}$ \Comment{Input: the output from residual block $N$}
\For{$\ell = N$ to $2$}
    \State {\small$(\vx^{\ell-1}, \vp^{\ell-1},\vn^{\ell-1})\gets \arg \min_{\vx,\vp,\vn}  \|\mathbf{y}^{\ell} - \mathbf{W}_{s}\mathbf{x} - \mathbf{W}_{2}\mathbf{p} \|^{2}_{2} + \lambda_{1}( \|\mathbf{n}^{T}\mathbf{p}\|_{2}^{2}) $ $+ \lambda_{2}( \|(\mathbf{W}_{1}\mathbf{x} - \mathbf{p} - \mathbf{n})\|)^{2}_{2}$}
    \State $\mathbf{y}^{\ell-1} = \vx^{\ell-1}$
\EndFor
\State $\Gamma(\vx) = \lVert \Phi( \mathbf{x}) - \widetilde{\Phi}_{0} \rVert^2 / \lVert \widetilde{\Phi}_{0} \rVert^2  + \lambda_{\alpha} R_{\alpha}(\mathbf{x}) + \lambda_{V\beta} R_{V\beta}(\mathbf{x}) $
\State ${\displaystyle \mathbf{x}^{*} \gets \arg \min_{\mathbf{x} \in \mathbb{R}^{H \times W \times C}} \Gamma(\vx)}$
\Ensure $\mathbf{x}^{*}$ \Comment{Output: reconstructed image}
\end{algorithmic}
}
\end{algorithm}

A typical ResNet architecture   (see Figure~\ref{PEEL_untrained})  has many Residual Blocks. For instance, a ResNet 18 architecture has 8 Residual Blocks. The series of Residual Blocks is typically preceded by shallow layers constituting convolution and maxpool layers (see Figure~\ref{resnet18}). Given an image representation taken from the output of a Residual Block, the goal of \textbf{PEEL} is to invert the preceding Residual Blocks and shallow layers to find the original input image. 

To accomplish this, \textbf{PEEL} iteratively solves Equation ~\eqref{resblock_inversion} to reconstruct the inputs of the Residual Blocks, starting at the deepest residual block and moving towards the start of the ResNet. Once the input to the first Residual Block in the network is recovered, \textbf{PEEL} then has to cope with the shallow layers consisting of convolutional layers and pooling. These shallow layers lose information, especially in the pooling layer, so are not invertible. However, the suppose the adversary knows that the inputs to the ResNets are natural images, so the embedding inversion approach of~\cite{2015_CVPR} is appropriate for inverting the shallow layers here. Thus, \textbf{PEEL} uses the reconstructed input $\widetilde{\Phi}_{0}$ to the last Residual block as input to the optimization problem in~\eqref{embedding_inv} to invert the shallow layers and obtain the desired approximation of the input image. 

The entire \textbf{PEEL} algorithm is given as Algorithm \ref{PEEL+embedding_inversion}. We note that \textbf{PEEL} is a generic and distribution-agnostic method that does not put any constraints on the similarity between the pre-train data for training ResNets and the user's private data to be reconstructed at inference time. Even when the weights of the ResNets are randomly initialized (i.e., untrained weights),  \textbf{PEEL} can still achieve effective data reconstruction.  \textbf{PEEL} also does not require the use of similar in-distribution data to the private data for data reconstruction.Figure~\ref{PEEL_untrained}~\ref{PEEL_trained} shows the empirical results in scenarios when the layers have a model with randomly initialized weights and when pre-trained weights (having a different distribution than test data)  are used for inversion. The results demonstrate the power of PEEL in reconstructing input data.

\section{Empirical Evaluation}

\label{sec:exp}
This section compares the performance of \textbf{PEEL} with state-of-the-art model inversion attacks within the commonly used context of model inversion on facial recognition systems. Since \textbf{Embedding Inversion} \cite{2015_CVPR} has been shown to perform poorly on deeper ResNets (e.g., see Figure \ref{fig:example}), we emphasize generative methods in our comparative analysis here.

We emphasize that \textbf{PEEL} is designed to recover inference-time inputs from the feature representations generated by the model, while the generative recovery methods--- ~\cite{GMI}, ~\cite{KEDMI},~\cite{VMI}, and \textbf{Rethink-MI}~\cite{rethink}--- are designed to recover variational approximations of the training data from the trained model itself. 
Nonetheless, these methods can be meaningfully compared by noting that in practice facial recognition systems are frequently employed to recognize the faces used in training the systems. When this is the case, \textbf{PEEL} will attempt to recover an image that is similar to the images used in training, similar to the goal of these generative approaches. The crucial difference is that \textbf{PEEL} aims to recover the specific input image, while generative methods aim to recover images that look similar to the true training data, in the sense that \emph{the generative model} thinks the images are similar. 

The work~\cite{rethink} provides several methodological improvements to these generative recovery methods and we consider the most competive baseline \textbf{KEDMI-LOMMA}~\cite{rethink} (an improvement on \textbf{KEDMI} \cite{KEDMI}) for a comparison of reconstruction (see Figure \ref{fig:reconstruction_comparisons}). 
For a fair comparison with these generative methods, we empirically ensure that the distributional recovery is converging. For a discussion on the training dynamics of these generative methods we refer to the Appendix \ref{sec: GAN_training}.




\subsection{Experimental Setup}

To evaluate the performance of \textbf{PEEL} on data recovery, we consider the IR-152 model (details in Appendix \ref{sec:IR-152}) \cite{rethink} and samples from the CelebA dataset \cite{celebA}. Unlike \textbf{PEEL}, generative methods are class-conditional, necessitating a subset of classes for a fair comparison. Specifically, we select 2 classes with 30 samples each (see Figure \ref{fig:reconstruction_comparisons}). Two standard metrics for reporting image reconstruction are Mean Squared Error (MSE) and Peak Signal-to-Noise Ratio (PSNR) \cite{jin2021cafe}.  For the generative method \textbf{KEDMI-LOMMA} (see experimental details in Appendix \ref{sec: GAN_training}), the MSE and PSNR are computed for the closest reconstruction of each sample in the target class. We consider different setups of the target model. \textbf{U}/\textbf{P} means target model uses \textbf{randomly initialized weights} /  \textbf{pretrained weights}, and \textbf{M} means presence of pooling.In the following experiments, unless otherwise noted , pretrained weights are from \textbf{ImageNet}.
Generative methods use KNN-Distance and Attack Accuracy \cite{rethink} as evaluation metrics. For a comprehensive analysis, the performance of \textbf{PEEL} on these metrics is presented in Appendix.

\begin{figure}[ht]
    \centering
    \begin{subfigure}[b]{0.23\textwidth} 
        \includegraphics[width=\textwidth]{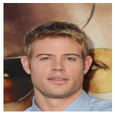}
        \caption{Original}
        \label{Original}
    \end{subfigure}
    \hspace{2mm} 
    \begin{subfigure}[b]{0.23\textwidth}
        \includegraphics[width=\textwidth]{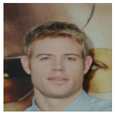}
        \caption{ResNet-34}
        \label{resnet-34}
    \end{subfigure}
    
    \begin{subfigure}[b]{0.23\textwidth}
        \includegraphics[width=\textwidth]{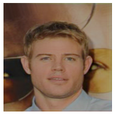}
        \caption{ResNet-50}
        \label{resnet-50}
    \end{subfigure}
    \hspace{2mm} 
    \begin{subfigure}[b]{0.23\textwidth}
        \includegraphics[width=\textwidth]{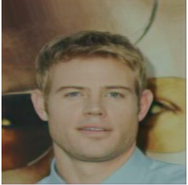}
        \caption{ResNet-152}
        \label{resnet-152}
    \end{subfigure}
    
    \vspace{-2mm} 
    \caption{PEEL is robust to deep layers in ResNets. (a) is the original image; (b)/(c)/(d) shows the reconstruction for ResNet-34/ResNet-50/ResNet-152.}
    \label{All_ResNets}
\end{figure}

\begin{figure}[ht]
    \centering
    
    \begin{subfigure}[b]{0.15\textwidth}
        \includegraphics[width=\textwidth]{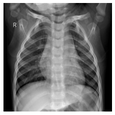}
        \caption{}
        \label{fig:normal}
    \end{subfigure}
    \hfill
    \begin{subfigure}[b]{0.15\textwidth}
        \includegraphics[width=\textwidth]{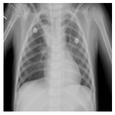}
        \caption{}
        \label{fig:bacterial}
    \end{subfigure}
    \hfill
    \begin{subfigure}[b]{0.15\textwidth}
        \includegraphics[width=\textwidth]{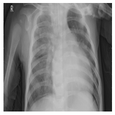}
        \caption{}
        \label{fig:viral}
    \end{subfigure} \\
    
    \begin{subfigure}[b]{0.15\textwidth}
        \includegraphics[width=\textwidth]{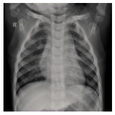}
        \caption{}
        \label{fig:normal_reconstructed}
    \end{subfigure}
    \hfill
    \begin{subfigure}[b]{0.15\textwidth}
        \includegraphics[width=\textwidth]{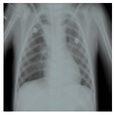}
        \caption{}
        \label{fig:bacterial_reconstructed}
    \end{subfigure}
    \hfill
    \begin{subfigure}[b]{0.15\textwidth}
        \includegraphics[width=\textwidth]{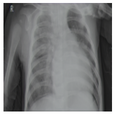}
        \caption{}
        \label{fig:viral_reconstructed}
    \end{subfigure}
    
    \vspace{-1mm}
    \caption{Reconstruction results for Chest X-ray images using ResNet-18 as the target model. (a) Normal Patient, (b) Bacterial Infection, (c) Viral Infection, (d) PEEL reconstruction of a Normal Patient, (e) PEEL reconstruction of a Bacterial Infection, (f) PEEL reconstruction of a Viral Infection.}
    \label{fig:viral_all}
    \vspace{-2mm}
\end{figure}

\begin{figure*}[t]
  \centering
  \begin{subfigure}[b]{0.17\textwidth}
    \includegraphics[width=\textwidth]{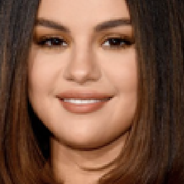}
    \caption{Input Image}
    \label{fig:original1}
  \end{subfigure}
  \begin{subfigure}[b]{0.17\textwidth}
    \includegraphics[width=\textwidth]{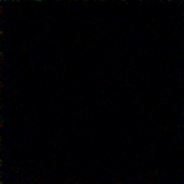}
    \caption{VGG-16}
    \label{fig:recovered1}
  \end{subfigure}
  \begin{subfigure}[b]{0.17\textwidth}
    \includegraphics[width=\textwidth]{Styles/pictures/recons_image_AllConv.png}
    \caption{VGG-19}
    \label{fig:recovered2}
  \end{subfigure}
  \begin{subfigure}[b]{0.17\textwidth}
    \includegraphics[width=\textwidth]{Styles/pictures/recons_image_AllConv.png}
    \caption{All Conv Net}
    \label{fig:recovered3}
  \end{subfigure}
  \begin{subfigure}[b]{0.17\textwidth}
    \includegraphics[width=\textwidth]{Styles/pictures/recons_image_AllConv.png}
    \caption{AlexNet}
    \label{fig:recovered4}
  \end{subfigure}

  \begin{subfigure}[b]{0.17\textwidth}
    \includegraphics[width=\textwidth]{Styles/pictures/input_image_shallow_1.png}
    \caption{Input Image}
    \label{fig:original2}
  \end{subfigure}
  \begin{subfigure}[b]{0.17\textwidth}
    \includegraphics[width=\textwidth]{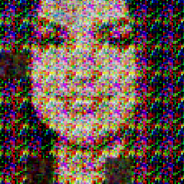}
    \caption{PEEL $N$ = 1}
    \label{fig:recovered5}
  \end{subfigure}
  \begin{subfigure}[b]{0.17\textwidth}
    \includegraphics[width=\textwidth]{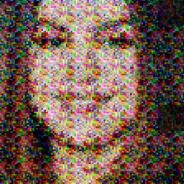}
    \caption{PEEL $N$ = 2}
    \label{fig:recovered6}
  \end{subfigure}
  \begin{subfigure}[b]{0.17\textwidth}
    \includegraphics[width=\textwidth]{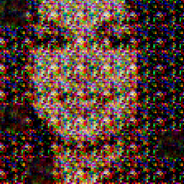}
    \caption{PEEL $N$ = 3}
    \label{fig:recovered7}
  \end{subfigure}
  \begin{subfigure}[b]{0.17\textwidth}
    \includegraphics[width=\textwidth]{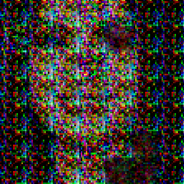}
    \caption{PEEL $N$ = 4}
    \label{fig:recovered8}
  \end{subfigure}
  \vspace{-1mm}
  \caption{The figures demonstrate the recovery limitations of \textbf{PEEL} in models without skip connections or residual propagation, as shown in \textbf{VGG-16/19} and \textbf{All Conv Net} (top row). Additionally, the effectiveness of \textbf{PEEL} in recovering inputs from \textbf{Vision Transformers} with various encoder lengths ($N$) is highlighted, showcasing different recovery qualities (bottom row).}
  \label{fig:res_important}
  \vspace{-2mm}
\end{figure*}

\begin{table}[ht]
\centering
\begin{tabular}{@{}lcc@{}}
\toprule
Method              & MSE            & PSNR          \\ \midrule
PEEL (\textbf{U})          & 774.17 ± 508.41        & 20.12 ± 2.91         \\
PEEL (\textbf{P})              & 3789.60 ± 1302.52      & 12.59 ± 1.51         \\
PEEL (\textbf{MU})  & 1916.53 ± 902.69       & 15.76 ± 2.06         \\
PEEL (\textbf{MP})     & 5770.21 ± 1730.52      & 10.71 ± 1.32         \\
Generative (KEDMI+LOMMA)  & 10493.42 ± 3837.30     & 8.28 ± 1.93          \\ 
Embedding Inversion & 18220.41± 1615.56     & 5.53 ± 1.44        \\ 
\bottomrule
\end{tabular}

\caption{Performance of different methods for class 1. There are 30 samples in each class and the MSE/PSNR is computed on the closest reconstruction for each sample in the class.  }
\label{tab:mse_class:1}
\vspace{-4mm}
\end{table}

\begin{table}[ht]
\centering
\begin{tabular}{@{}lcc@{}}
\toprule
Method              & MSE              & PSNR          \\ \midrule
PEEL (\textbf{U})         & 402.99 ± 297.74        & 23.24 ± 3.36    \\
PEEL (\textbf{P})             & 2603.77 ± 1009.05      & 14.29 ± 1.70         \\
PEEL (\textbf{MU}) & 1070.85 ± 621.33       & 18.55 ± 2.59         \\

PEEL (\textbf{MP})     & 4151.99 ± 1493.58      & 12.21 ± 1.55         \\
Generative (KEDMI + LOMMA)  & 10149.92 ± 2430.76     & 8.18 ± 1.02  \\ 
Embedding Inversion & 10534.67± 1834.22    & 7.90 ± 1.73       \\         
\bottomrule
\end{tabular}
\caption{Performance of different methods for class 2. There are 30 samples in each class and the MSE /PSNR is computed on the closest reconstruction for each sample in the class }
\label{tab:mse_class:2}
\end{table}

\begin{figure*}[t]
\centering
\begin{tabular}{c c c c}
    \begin{subfigure}[t]{0.24\textwidth}
        \centering
        \includegraphics[width=\textwidth]{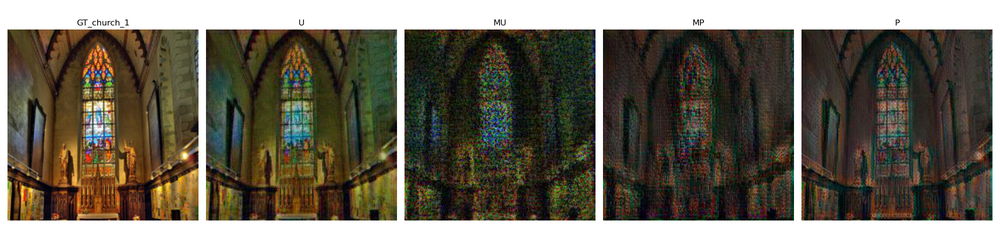}\\
        \includegraphics[width=\textwidth]{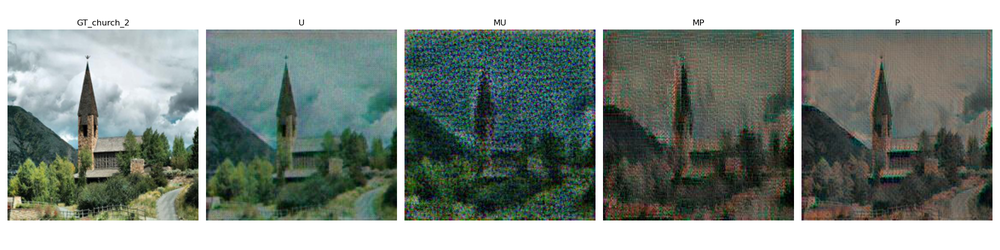}\\
        \includegraphics[width=\textwidth]{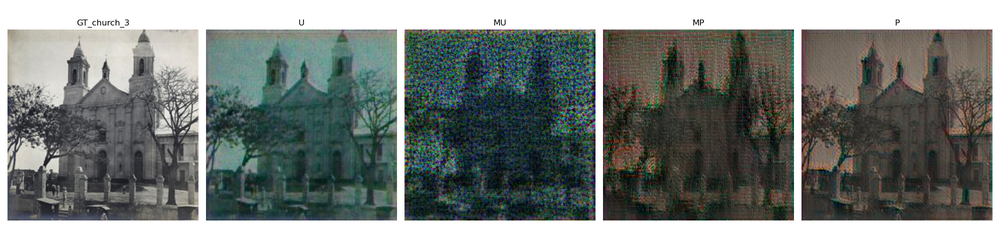}

        \vspace{4pt}
        \caption{Church}
    \end{subfigure}
    &
    \begin{subfigure}[t]{0.24\textwidth}
        \centering
        \includegraphics[width=\textwidth]{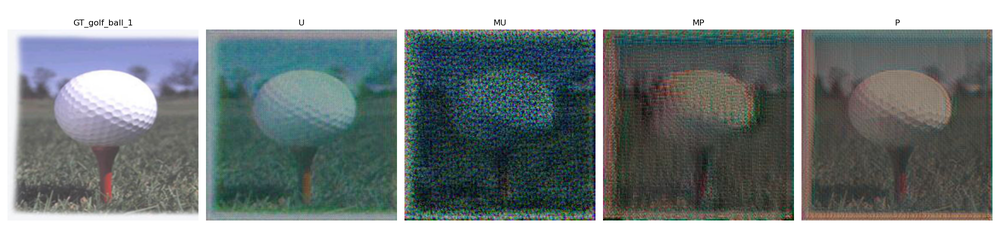}\\
        \includegraphics[width=\textwidth]{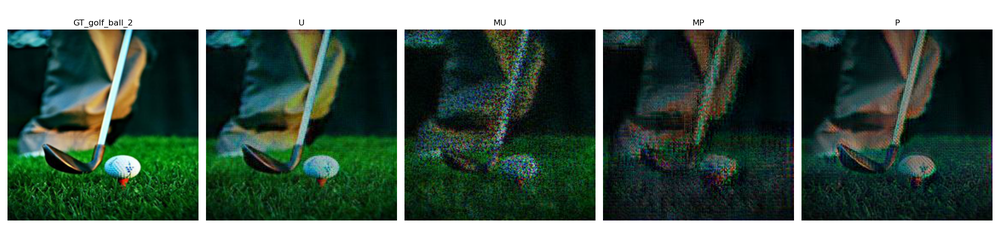}\\
        \includegraphics[width=\textwidth]{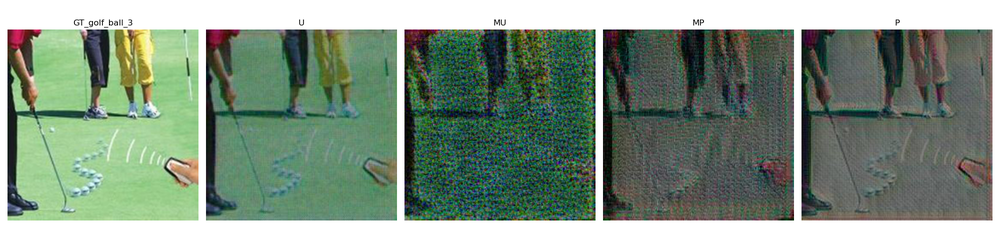}

        \vspace{4pt}
        \caption{Golf Ball}
    \end{subfigure}
    &
    \begin{subfigure}[t]{0.24\textwidth}
        \centering
        \includegraphics[width=\textwidth]{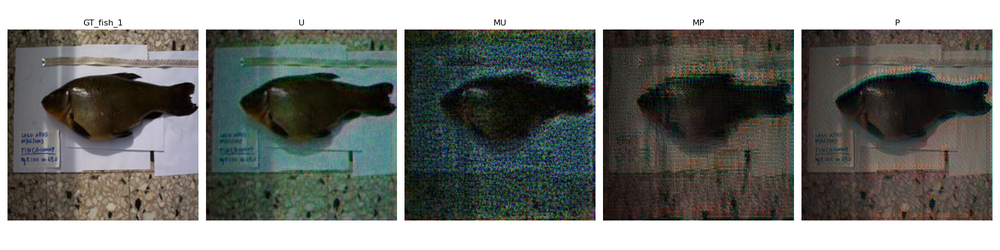}\\
        \includegraphics[width=\textwidth]{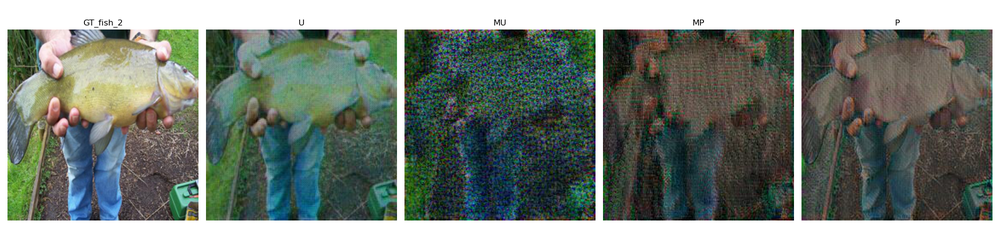}\\
        \includegraphics[width=\textwidth]{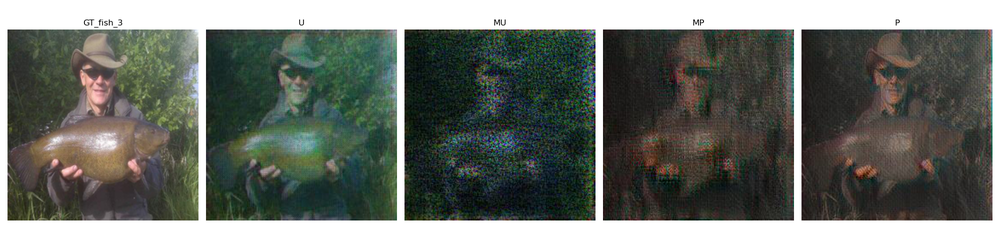}

        \vspace{4pt}
        \caption{Fish}
    \end{subfigure}
    &
    \begin{subfigure}[t]{0.24\textwidth}
        \centering
        \includegraphics[width=\textwidth]{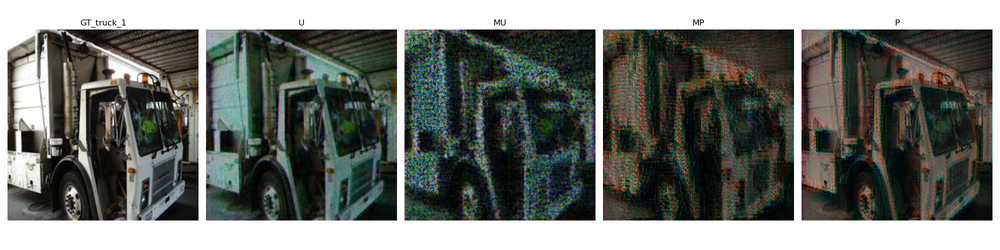}\\
        \includegraphics[width=\textwidth]{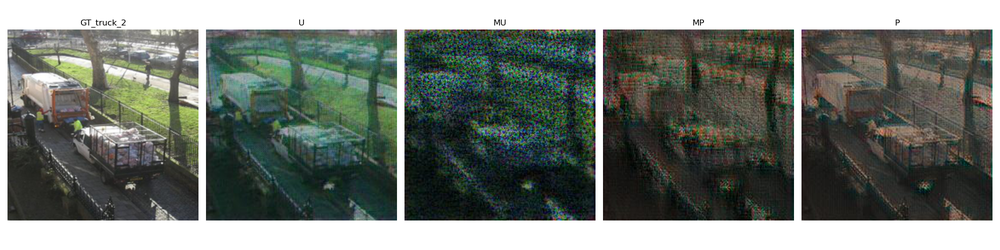}\\
        \includegraphics[width=\textwidth]{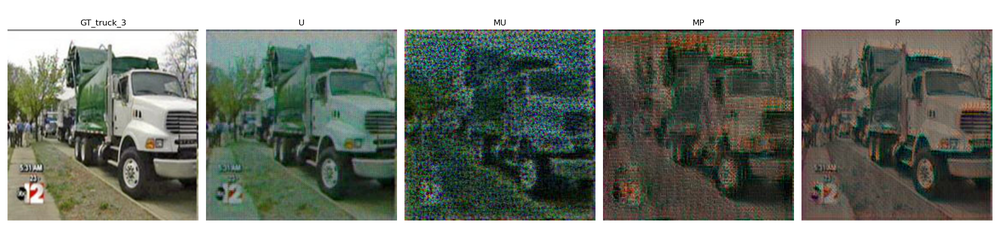}

        \vspace{4pt}
        \caption{Truck}
    \end{subfigure}
\end{tabular}

\caption{Four ImageNet classes are presented. Each column contains three samples from each class with different recovery configurations (from top to bottom). The underlying model is a ResNet-18 trained on ImageNet (for the \textbf{P} variant of \textbf{PEEL}) . Similar to Figure \ref{fig:reconstruction_comparisons}, the order for each sample is:\textbf{ GROUND TRUTH, PEEL (U), PEEL (MU), PEEL (P), and PEEL (MP)}.}
\label{fig:imagenet_comparisons}
\end{figure*}

\begin{figure*}[t]
\centering
\begin{tabular}{c c c c}
    \begin{subfigure}[t]{0.24\textwidth}
        \centering
        \includegraphics[width=\textwidth]{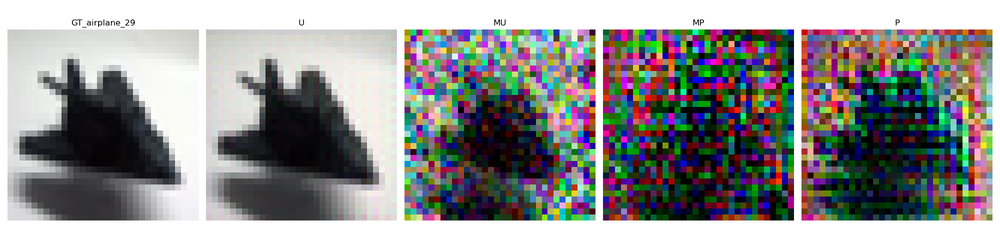}\\
        \includegraphics[width=\textwidth]{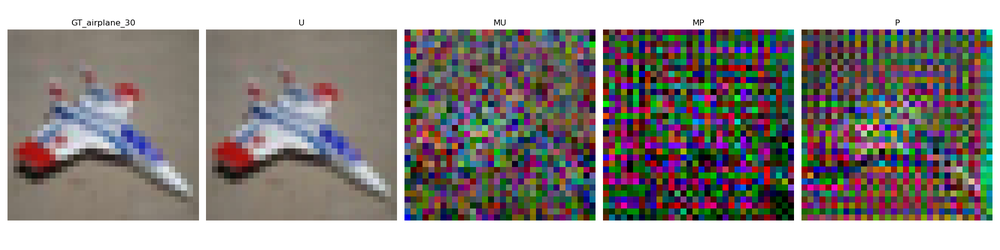}\\
        \includegraphics[width=\textwidth]{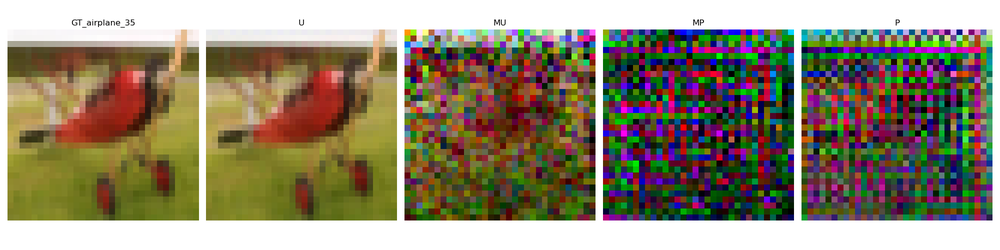}

        \vspace{4pt}
        \caption{Airplane}
    \end{subfigure}
    &
    \begin{subfigure}[t]{0.24\textwidth}
        \centering
        \includegraphics[width=\textwidth]{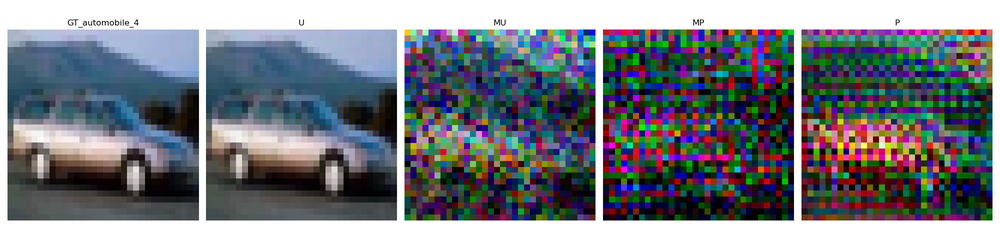}\\
        \includegraphics[width=\textwidth]{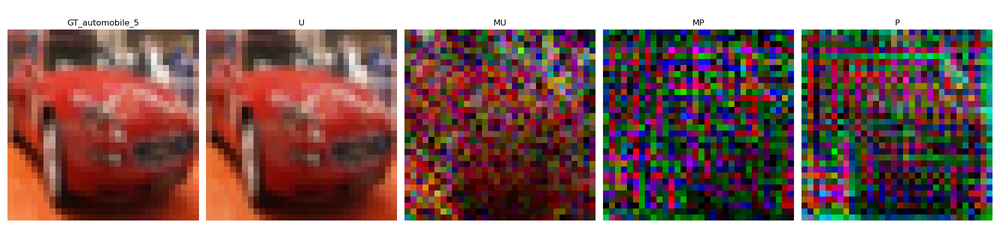}\\
        \includegraphics[width=\textwidth]{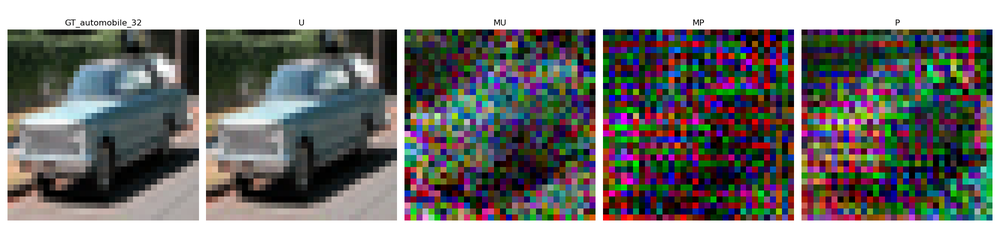}

        \vspace{4pt}
        \caption{Automobile}
    \end{subfigure}
    &
    \begin{subfigure}[t]{0.24\textwidth}
        \centering
        \includegraphics[width=\textwidth]{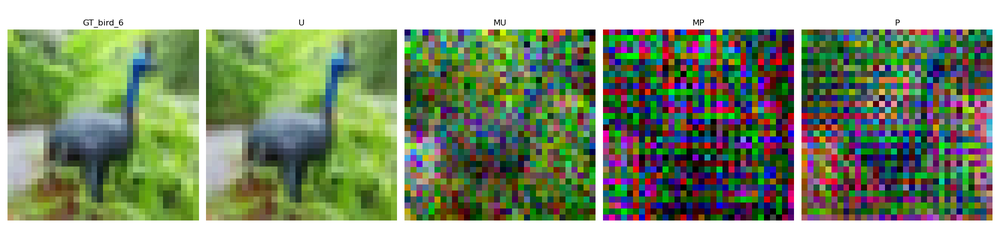}\\
        \includegraphics[width=\textwidth]{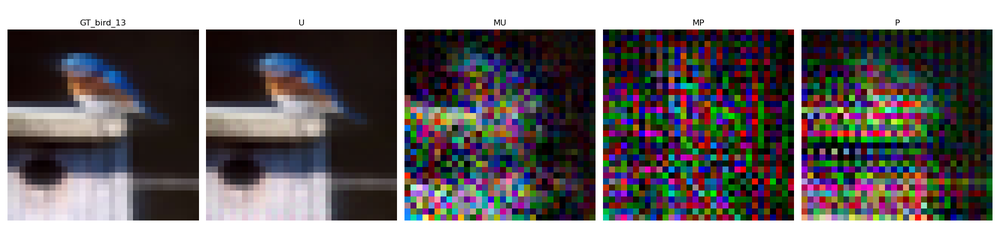}\\
        \includegraphics[width=\textwidth]{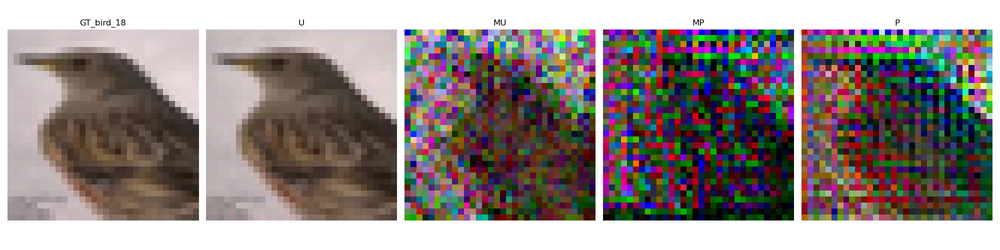}

        \vspace{4pt}
        \caption{Bird}
    \end{subfigure}
    &
    \begin{subfigure}[t]{0.24\textwidth}
        \centering
        \includegraphics[width=\textwidth]{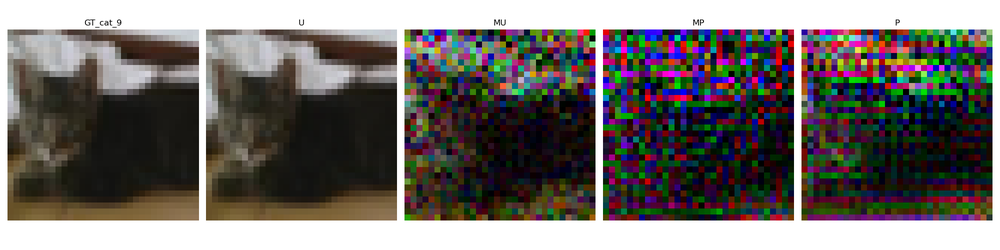}\\
        \includegraphics[width=\textwidth]{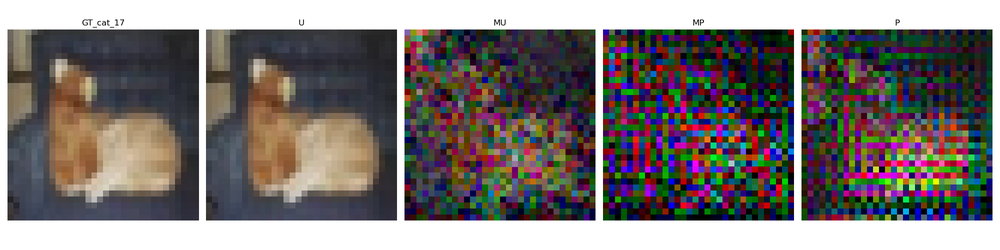}\\
        \includegraphics[width=\textwidth]{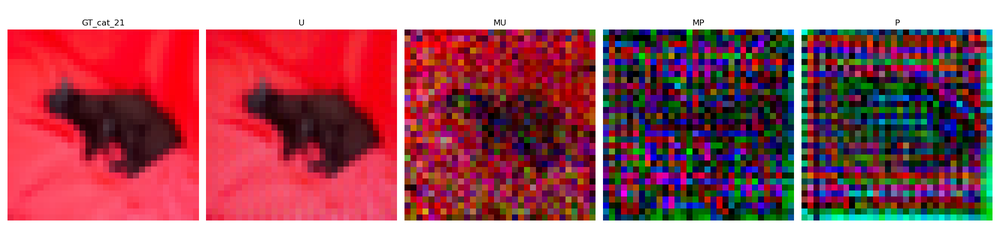}

        \vspace{4pt}
        \caption{Cat}
    \end{subfigure}
\end{tabular}

\caption{Four CIFAR-10 classes are presented. Each column contains three samples from each class with different recovery configurations (from top to bottom). The underlying model is a ResNet-18 trained on ImageNet. Similar to Figure \ref{fig:reconstruction_comparisons}, the order for each sample is:\textbf{ GROUND TRUTH, PEEL (U), PEEL (MU), PEEL (P), and PEEL (MP)}.}
\label{fig:cifar_comparisons}
\end{figure*}


\begin{figure*}[t]
\centering
\begin{tabular}{c c}
    \scriptsize \textbf{Bacteria} & 
    \scriptsize \textbf{Virus} \\
    \includegraphics[width=0.5\textwidth]{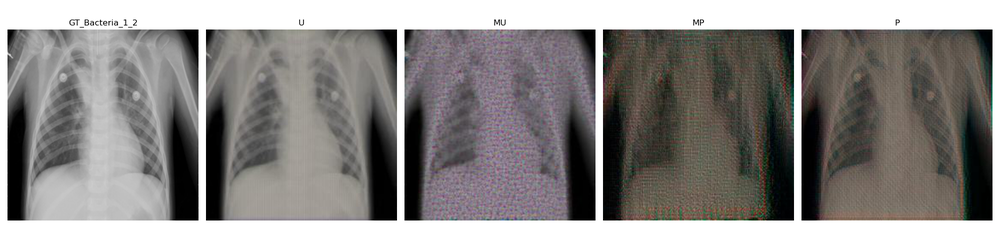} &
    \includegraphics[width=0.5\textwidth]{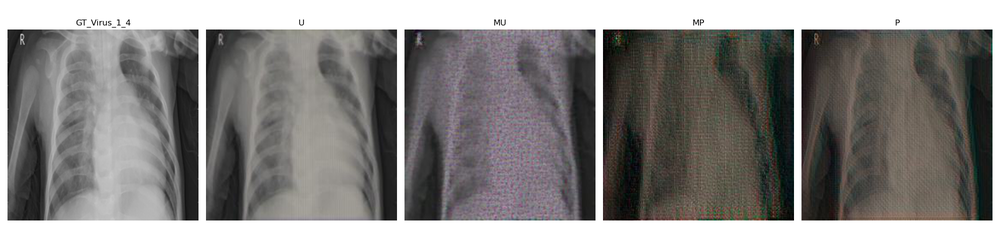} \\
    
    \scriptsize \textbf{Normal 1} & 
    \scriptsize \textbf{Normal 2} \\
    \includegraphics[width=0.5\textwidth]{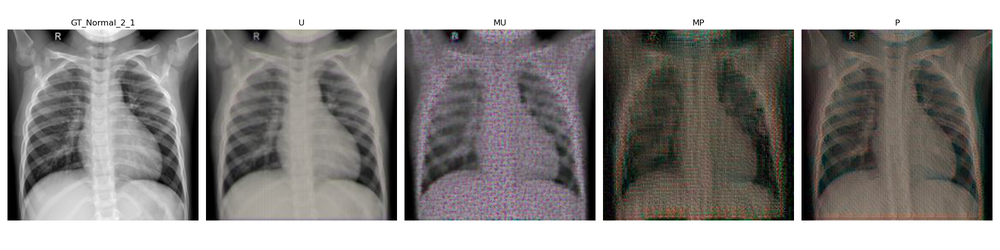} &
    \includegraphics[width=0.5\textwidth]{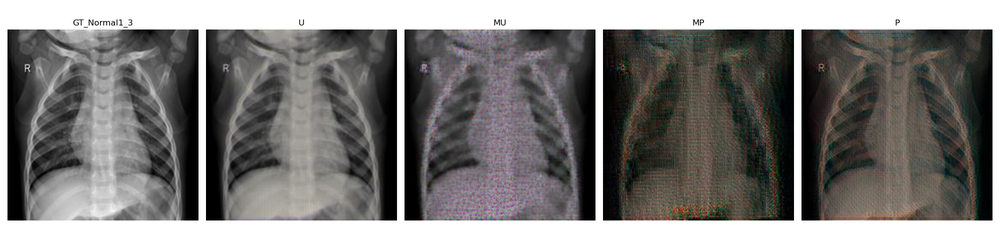} \\
\end{tabular}
\caption{This figure compares different configurations of models for chest X-ray images (Bacteria, Virus, Normal~1, Normal~2). 
Each image (left to right within it) represents a particular configuration: 
\textbf{U} (randomly initialized weights), 
\textbf{MU} (pooling + randomly initialized), 
\textbf{MP} (pooling + pretrained weights), 
\textbf{P} (pretrained weights only). 
All samples use ResNet-18. 
\textbf{PEEL} generally achieves robust recovery across classes, but the chosen regularization can affect low-resolution images differently.}
\label{fig:chest_configuration}
\end{figure*}

\begin{figure*}[t]
\centering
\begin{tabular}{c c}
  \scriptsize \textbf{Bacteria} & 
  \scriptsize \textbf{Virus} \\
  \includegraphics[width=0.5\textwidth]{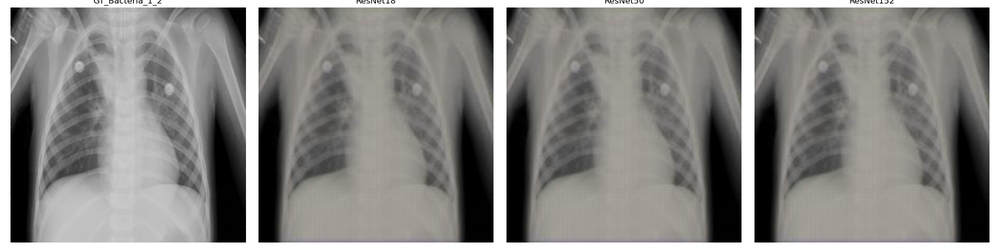} &
  \includegraphics[width=0.5\textwidth]{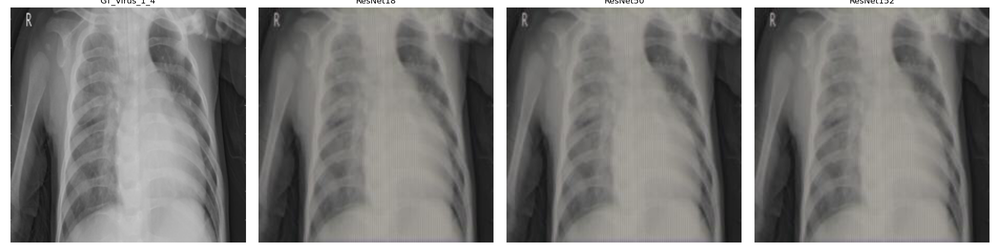} \\
  
  \scriptsize \textbf{Normal 1} &
  \scriptsize \textbf{Normal 2} \\
  \includegraphics[width=0.5\textwidth]{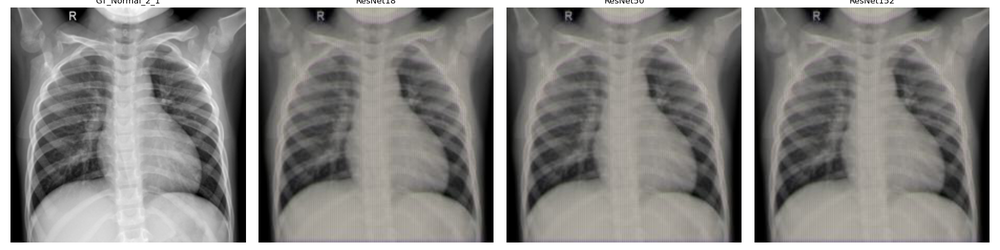} &
  \includegraphics[width=0.5\textwidth]{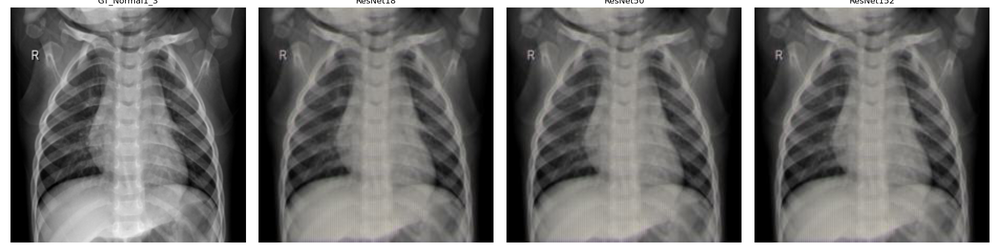} \\
\end{tabular}

\caption{This figure compares different models for chest X-ray image classification across varying model depths. 
Each image shows the ground truth on the left, then results from ResNet-18, ResNet-50, and ResNet-152 (from left to right) 
for each class: \textbf{Bacteria}, \textbf{Virus}, \textbf{Normal 1}, and \textbf{Normal 2}. 
Similar to the results in Figure~\ref{All_ResNets}, all tested depths yield consistently high-quality input recovery, 
indicating that deeper networks (e.g., ResNet-152) do not compromise recovery performance for \textbf{PEEL}.}
\label{fig:chest_model_depth}
\end{figure*}

\begin{figure*}
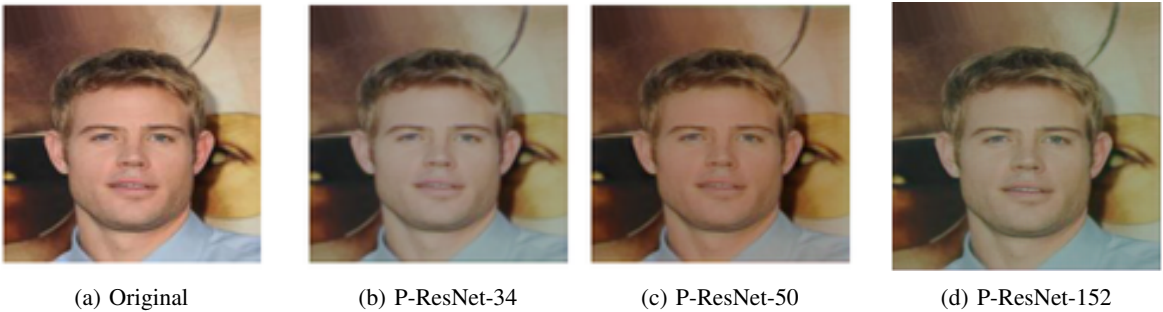

    \centering
    \begin{subfigure}[b]{0.20\textwidth} 
        \includegraphics[width=\textwidth]{Styles/pictures/CelebA_orignal.png}
        \caption{Original}
        \label{Original}
    \end{subfigure}
    \hspace{2mm} 
    \begin{subfigure}[b]{0.20\textwidth}
        \includegraphics[width=\textwidth]{Styles/pictures/Resnet-34.png}
        \caption{P-ResNet-34}
        \label{resnet-34}
    \end{subfigure}
    \begin{subfigure}[b]{0.20\textwidth}
        \includegraphics[width=\textwidth]{Styles/pictures/resnet50.png}
        \caption{P-ResNet-50}
        \label{resnet-50}
    \end{subfigure}
    \hspace{2mm} 
    \begin{subfigure}[b]{0.20\textwidth}
        \includegraphics[width=\textwidth]{Styles/pictures/ResNet152.png}
        \caption{P-ResNet-152}
        \label{resnet-152}
    \end{subfigure}
    
    \vspace{-2mm} 
    \caption{\textbf{PEEL} demonstrates robustness to deeper layers in P-ResNets (where "P-" indicates the use of PReLU as the activation function). In the figure, (a) represents the original image, while (b), (c), and (d) show the reconstructions using P-ResNet-34, P-ResNet-50, and ResNet-152, respectively, where the standard ReLU activation has been replaced with PReLU.}
    \label{fig: PResNet}
\end{figure*}

\begin{figure*}
    \centering

    
    \begin{subfigure}[b]{0.15\textwidth}
        \centering
        \includegraphics[width=\textwidth]{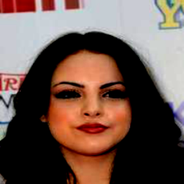}
    \end{subfigure}
    \begin{subfigure}[b]{0.15\textwidth}
        \centering
        \includegraphics[width=\textwidth]{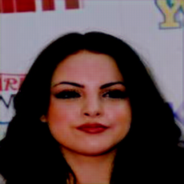}
    \end{subfigure}
    \begin{subfigure}[b]{0.15\textwidth}
        \centering
        \includegraphics[width=\textwidth]{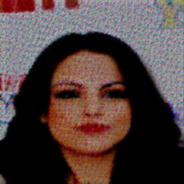}
    \end{subfigure}
    \begin{subfigure}[b]{0.15\textwidth}
        \centering
        \includegraphics[width=\textwidth]{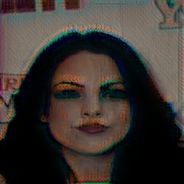}
    \end{subfigure}
    \begin{subfigure}[b]{0.15\textwidth}
        \centering
        \includegraphics[width=\textwidth]{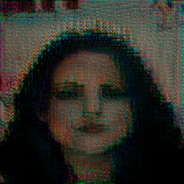}
    \end{subfigure}
    
    \vspace{1em}
    \begin{subfigure}[b]{0.15\textwidth}
        \centering
        \includegraphics[width=\textwidth]{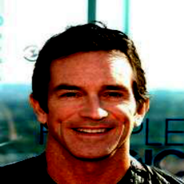}
    \end{subfigure}
    \begin{subfigure}[b]{0.15\textwidth}
        \centering
        \includegraphics[width=\textwidth]{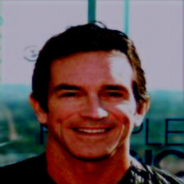}
    \end{subfigure}
    \begin{subfigure}[b]{0.15\textwidth}
        \centering
        \includegraphics[width=\textwidth]{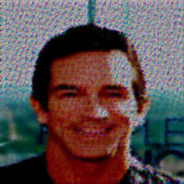}
    \end{subfigure}
    \begin{subfigure}[b]{0.15\textwidth}
        \centering
        \includegraphics[width=\textwidth]{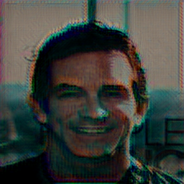}
    \end{subfigure}
    \begin{subfigure}[b]{0.15\textwidth}
        \centering
        \includegraphics[width=\textwidth]{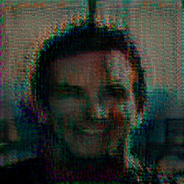}
    \end{subfigure}
    
    \vspace{1em}
    \begin{subfigure}[b]{0.15\textwidth}
        \centering
        \includegraphics[width=\textwidth]{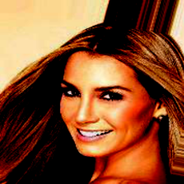}
    \end{subfigure}
    \begin{subfigure}[b]{0.15\textwidth}
        \centering
        \includegraphics[width=\textwidth]{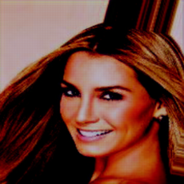}
    \end{subfigure}
    \begin{subfigure}[b]{0.15\textwidth}
        \centering
        \includegraphics[width=\textwidth]{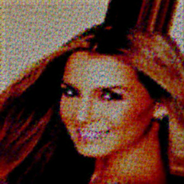}
    \end{subfigure}
    \begin{subfigure}[b]{0.15\textwidth}
        \centering
        \includegraphics[width=\textwidth]{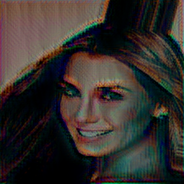}
    \end{subfigure}
    \begin{subfigure}[b]{0.15\textwidth}
        \centering
        \includegraphics[width=\textwidth]{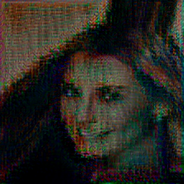}
    \end{subfigure}
    
    \caption{The figure presents samples from the CelebA dataset, with the ground truth images on the far left, followed by reconstructions under various configurations. From left to right, the reconstructions represent the following scenarios: randomly initialized weights, pooling only, pretrained weights, and pooling with pretrained weights.The results across different classes demonstrate high-quality input recovery for each configuration, highlighting the robustness of \textbf{PEEL} in effectively recovering inputs across samples.}
    \label{fig:add_celebA_class_comparisons}
\end{figure*}

\textbf{Hyperparameters:} used for training the target and evaluation models were adopted from \cite{rethink} to ensure a fair comparison with the performance metrics reported in that work. For PEEL, a fixed penalty was used to recover the inputs of residual blocks: the parameters $\lambda_{1}$ and $\lambda_{2}$ in Equation \eqref{penalty_equation} were set to 1000. The choice of these hyperparameters is further discussed in the Appendix. The Adam optimizer was employed with a constant learning rate of $\eta=0.01$, and the number of epochs used to solve Equation \eqref{penalty_equation} for each residual block was set to 2000.

\subsection{Performance Comparison}

The reconstruction results are presented in Tables \ref{tab:mse_class:1} and \ref{tab:mse_class:2}. 
\textbf{PEEL} achieves high-quality reconstruction of images across different classes and variants. However, in the presence of non-invertible pooling layers, the reconstruction quality suffers slightly. Additionally, since \textbf{PEEL} is a numerical solver, the quality of the reconstruction decreases when large pretrained weights are used for recovery. Despite this, \textbf{PEEL} is robust to Residual Networks of varying depths, as seen in Figure \ref{All_ResNets}. Additionally, as discussed, the performance of \textbf{PEEL} is not dependent on the input data distribution. We evaluate its performance on the Chest X-ray dataset \cite{VMI}. \textbf{PEEL} maintains high-quality reconstruction even on non-facial datasets, as shown in Figure \ref{fig:viral_all}. 

Despite superior recovery performance, we also discuss some limitations observed in \textbf{PEEL}. \textbf{PEEL} is dependent on the presence of residual connections in the target model (see Figure \ref{fig:res_important}).  We conduct experiments on \textbf{Vision Transformers} \cite{visionT} and other deep neural networks without residual connections, such as \textbf{VGG} \cite{VGG}, \textbf{AllConvNets} \cite{springenberg2014striving}, and \textbf{AlexNet} \cite{alex}, to illustrate the importance of residual connections. 
For \textbf{Vision Transformers}, the presence of residual connections allows \textbf{PEEL} to reconstruct images by accessing outputs from different encoders. In the absence of such residual connections, \textbf{PEEL} fails to reconstruct the input image as shown by \textbf{PEEL}'s performance on non-residual architectures such as \textbf{VGG} \cite{VGG}, \textbf{AllConvNets} \cite{springenberg2014striving}, and \textbf{AlexNet} 
\cite{alex}.

\begin{table}[t]
\centering
\begin{tabular}{l l c c}
\toprule
\textbf{Layer} & \textbf{Method} & \textbf{MSE} & \textbf{PSNR} \\
               &                 & \textbf{(mean ± std)} & \textbf{(mean ± std)} \\
\midrule
Layer 1 & HRN-5x5x3x & 1798.62 ± 899.72 & 16.02 ± 2.60 \\
Layer 1 & HRN-2x5x3x & 1698.53 ± 852.94 & 16.48 ± 2.62 \\
Layer 1 & ResNet-18 & 1690.03 ± 890.24 & 16.50 ± 2.60 \\
\midrule
Layer 2 & HRN-5x5x3x & 1415.74 ± 703.64 & 18.17 ± 2.56 \\
Layer 2 & HRN-2x5x3x & 1259.82 ± 623.76 & 18.62 ± 2.56 \\
Layer 2 & ResNet-18 & 1400.12 ± 700.47 & 18.50 ± 2.55 \\
\midrule
Layer 3 & HRN-5x5x3x & 437.29 ± 205.64 & 23.09 ± 2.36 \\
Layer 3 & HRN-2x5x3x & 443.87 ± 206.72 & 23.02 ± 2.30 \\
Layer 3 & ResNet-18 & 430.01 ± 200.34 & 23.50 ± 2.35 \\
\midrule
Layer 4 & HRN-5x5x3x & 210.27 ± 68.89 & 25.83 ± 1.45 \\
Layer 4 & HRN-2x5x3x & 216.30 ± 69.48 & 25.66 ± 1.45 \\
Layer 4 & ResNet-18 & 205.50 ± 65.29 & 26.00 ± 1.45 \\
\bottomrule
\end{tabular}
\caption{Layer-wise reconstruction results for architectures with different filter widths are shown. Different filter widths control the distribution of non-linearities in the architecture. The performance differences between the architectures, measured by MSE and PSNR, are less than 1.5\% across all four layers. This indicates that despite variations in the number of nonlinearities within the residual blocks, the skip connections play a crucial role in providing residual information, enabling \textbf{PEEL} to recover the input with consistently high quality }
\label{tab:filter_widths}
\end{table}

\section{Ablation Study and  Additional Discussion on PEEL} 

\subsection{Further Experiments on Chest X Ray Dataset}

To provide a thorough evaluation of \textbf{PEEL} on the Chest X-ray dataset, we present results under various configurations. Figure \ref{fig:chest_configuration} illustrates the impact of different pretraining levels and the application of pooling layers, similar to the comparisons in Figure \ref{fig:reconstruction_comparisons}. The experiments show that, as with facial image recovery, \textbf{PEEL} is capable of high-resolution recovery of Chest X-ray samples under different configurations of the target model. However, the use of heavily pretrained weights and pooling layers in the target model can affect the quality of the recovered images due to its numerical limitations as discussed previously.

Furthermore, Figure \ref{fig:chest_model_depth} demonstrates the performance of \textbf{PEEL} across various model depths, from ResNet-18 to ResNet-152, analogous to the results in Figure \ref{All_ResNets}. \textbf{PEEL} consistently achieves strong performance across different model depths when applied to Chest X-ray samples, confirming its robustness regardless of the depth of the model used.

\subsection{PEEL on CIFAR and ImageNet}

We also evaluate \textbf{PEEL} on images from standard computer vision datasets such as \textbf{ImageNet} and \textbf{CIFAR}, as shown in Figure \ref{fig:imagenet_comparisons}. In these experiments, \textbf{PEEL} and its variants were tested across a subset of ImageNet classes. Visual inspection reveals that \textbf{PEEL} achieves recovery performance comparable to that observed for facial  samples (see Figure \ref{fig:reconstruction_comparisons}), demonstrating its ability to recover input images with high resolution for different training distributions.

Similarly, reconstruction results for CIFAR-10, presented in Figure \ref{fig:cifar_comparisons}, highlight \textbf{PEEL}'s performance on low-resolution images. While \textbf{PEEL} demonstrates strong recovery with randomly initialized weights, certain variants exhibit reduced effectiveness. This reduction is primarily due to the influence of the smoothness regularizer during the final recovery stage, which can impact the reconstruction of low-resolution CIFAR-10 images. Additional empirical results showing \textbf{PEEL}'s performance on other CIFAR-10 and ImageNet classes are discussed in the Appendix

\subsection{PEEL on more classes from Celeb A }

As discussed in previous sections, \textbf{PEEL} operates independently of the class-conditional training distribution of the target model, unlike generative methods that rely on such distributions. To further illustrate this, Figure \ref{fig:add_celebA_class_comparisons} presents additional examples from various classes, demonstrating the performance of \textbf{PEEL} across different target model configurations, including  pretraining and pooling layers. The results show that the performance of \textbf{PEEL} remains consistent across samples from different classes, indicating its robustness regardless of the class distribution.

\subsection{PEEL with linear Activation}
To emphasize the importance of the residual connection in \textbf{PEEL}'s input recovery, we conducted experiments by removing the non-linearity from the activation function and observed \textbf{PEEL}'s behavior with a linear activation function, specifically the Parametric Rectified Linear Unit (PReLU). PReLU \cite{thakur2020prelu} is a variant of the ReLU activation function in which the negative slope is learned during training. Mathematically, it is defined as:

\begin{align*}
PReLU(x) = 
\begin{cases} 
x, & \text{if } x \geq 0, \\
\alpha x, & \text{if } x < 0,
\end{cases}
\end{align*}

where \( \alpha \) is a learnable parameter. Unlike ReLU, which sets negative inputs to zero, PReLU scales negative values according to the learned parameter \( \alpha \). Figure \ref{fig: PResNet} illustrates the results, showing a comparable performance to Figure \ref{All_ResNets}, where a non-linear ReLU activation is used.

\subsection{PEEL with Changing Filter Widths}

We also evaluate \textbf{PEEL} in scenarios where the distribution of non-linearities is non-uniform. \cite{jha2023deepreshape} explore such ResNet variants by altering filter widths. To assess the impact of \textbf{PEEL} when the distribution of non-linearities varies across layers, we conducted experiments using the HybReNets architecture series, specifically the HRN-5x5x3x and HRN-2x5x3x models from \cite{jha2023deepreshape}, and compared their performance to the baseline ResNet-18 model.

In the table \ref{tab:filter_widths}, we present layer-wise reconstruction results for the three models. We sampled 10 different images from various classes within the CelebA dataset. The table displays the mean and standard deviation of the reconstruction outcomes after each layer. Layer 1 represents the first set of residual blocks, and Layer 4 corresponds to the final set. It’s important to note that ResNet-18 comprises four layers, each containing two residual blocks. Additionally, we have omitted the pooling layers in this analysis to specifically assess the impact of the distribution of non-linearity on the reconstruction results.

The performance differences for the various architectures, in terms of MSE and PSNR, are less than 1.5\% for each of the four layers. Thus, despite the difference in the number of nonlinearities within the residual blocks, the skip connections contribute residual information that aids PEEL in effectively recovering the input with high quality.

\subsection{Why PEEL Works Well on Residual Networks ?}

\textbf{PEEL }succeeds with ResNets because the skip connection adds the input $\mathbf{x}$ directly to a transformed version of $\mathbf{x}$ (i.e., $\mathbf{W}_2 \mathbf{p}$). This creates a near-linear relationship between the residual block’s output $\mathbf{y}$ and its input $\mathbf{x}$:
\[
  \mathbf{y} \;=\; \mathbf{W}_s \mathbf{x} \;+\; \mathbf{W}_2 \,\mathrm{ReLU}(\mathbf{W}_1 \mathbf{x}),
\]
where $\mathbf{W}_s$ may be an identity or downsampling convolution. Such skip connections effectively preserve important information about $\mathbf{x}$ in $\mathbf{y}$, making inversion more tractable. By contrast, purely feedforward layers without skip connections introduce cascades of non-linearities (e.g., ReLU) that obscure the original input, thereby hindering inversion.

\subsection{Pretrained Weights vs.\ Random Initialization}

\subsubsection*{(A) Darker Reconstructions with Pretrained Weights}
Experimental results (e.g., Figures~\ref{fig:reconstruction_comparisons},\ref{fig:imagenet_comparisons}--\ref{fig:chest_configuration} and \ref{fig:add_celebA_class_comparisons}) show that reconstructions from pretrained networks tend to be ``darker.'' We hypothesize two main factors:

\begin{enumerate}
  \item \textbf{Penalization Effects.}
    Our objective (see Equation \ref{penalty_equation}) includes a penalty term (e.g., $\mathbf{n}^\top \mathbf{p}$) which encourages smaller magnitudes for certain latent variables. When large pretrained weights amplify the network’s responses, the optimizer may favor solutions with suppressed pixel values in $\mathbf{x}$, thus yielding darker images. 
    Balancing this penalty with adaptive regularization constants could mitigate such issues.

  \item \textbf{Sensitivity of Pretrained Networks.}
    Networks trained on large datasets (e.g., ImageNet) can become ill-conditioned, exhibiting high sensitivity to input perturbations~\cite{References5,References4}. This yields a steep, unstable optimization landscape during inversion, causing gradients to explode or vanish. Consequently, the solution often converges to lower-intensity reconstructions, especially if the weight matrices have large spectral norms. Techniques like weight normalization or additional pixel-intensity priors (akin to the ``dark channel prior'' in~\cite{References3}) might help correct this bias.
\end{enumerate}

\subsubsection*{(B) Why Random Weights Often Yield Better Reconstructions} 

As seen in figures~\ref{fig:reconstruction_comparisons},\ref{fig:imagenet_comparisons}--\ref{fig:chest_configuration} and \ref{fig:add_celebA_class_comparisons}) randomly initialized weights (e.g., Xavier or Gaussian initialization) are typically better-conditioned than heavily trained weights, making them less sensitive to small changes in $\mathbf{x}$. During \textbf{PEEL}’s pseudo-inverse step
\[
  \mathbf{W}\,\mathbf{x} \;=\; \mathbf{p} \;-\; \mathbf{n},
\]
these well-conditioned matrices lead to more stable solutions, preserving pixel intensities and reducing numerical errors. As a result, when ResNet weights are random (i.e., untrained), \textbf{PEEL }often reconstructs images more faithfully.

\subsection*{IID vs.\ OOD Evaluation}

Figures~\ref{fig:reconstruction_comparisons}, \ref{fig:imagenet_comparisons}--\ref{fig:chest_configuration}, and \ref{fig:add_celebA_class_comparisons} show PEEL's performance on both in-distribution (IID) and out-of-distribution (OOD) samples to evaluate robustness. In IID experiments, where the target model (ResNet-18) was trained on ImageNet and queried with ImageNet test data, \textbf{PEEL} consistently recovered high-fidelity reconstructions (see Figure~\ref{fig:imagenet_comparisons}). Under OOD settings—including Chest X-ray images, CelebA facial images, and the low-resolution CIFAR-10 dataset—PEEL still achieved strong performance in terms of reconstruction quality (Figures~\ref{fig:cifar_comparisons}, \ref{fig:add_celebA_class_comparisons}, \ref{fig:chest_configuration}), indicating that ResNet skip connections preserve sufficient information for inversion even across sizable distributional gaps.

\section{Conclusion}
\label{sec:conc}
Residual blocks form the backbone of many deep learning architectures, including ResNets and transformers. Ensuring the privacy of inference data is crucial for model trustworthiness, making it imperative to study these architectures for potential data leakage during inference. In this work, we proposed a novel method, \textbf{PEEL}, which employs an advanced embedding inversion approach to conduct inference-time data leakage attacks on residual block architectures. The empirical success of \textbf{PEEL} validates the intuition that residual blocks output transformed versions of their inputs that can, in practice, be inverted. Despite involving non-convex optimization, residual blocks are sufficiently susceptible to leakage, allowing inversion even in deep residual networks.

We conducted experiments using \textbf{PEEL} on samples from diverse distributions, including facial recognition data, chest X-rays, and standard datasets like ImageNet and CIFAR-10. We evaluated \textbf{PEEL} under various settings by altering the depth of ResNet models—from ResNet-18 to ResNet-152—and adjusting model widths through filter distributions and non-linearities. Both linear and non-linear activations were considered. Our results demonstrate that \textbf{PEEL} achieves high recovery quality across all settings, although non-invertible pooling layers and the use of pre-trained weights can affect the effectiveness of recovery. To emphasize the importance of residual connections in recovering intermediate representations, we also experimented with vision transformers that utilize skip connections, as well as other deep networks like AllConvNets \cite{springenberg2014striving} and AlexNet \cite{alex}. These experiments confirm the necessity of residual connections for high-quality recovery using \textbf{PEEL}, thus raising important privacy concerns regarding the use of residual architectures where potential  $\vH\vb\vC$  adversaries may infer input inference data.

\textbf{PEEL} leverages the linear component of skip connections to effectively invert deep networks, demonstrating consistent performance across diverse distributions. Notably, \textbf{PEEL} is particularly well-conditioned when applied to networks with random rather than pretrained weights. Furthermore, \textbf{PEEL} can be extended to other residual architectures, as discussed briefly in Appendix~\ref{ref:adapt_new_PEEL}. Future efforts may explore advanced regularization or weight-normalization techniques to enhance visual fidelity, especially in the presence of large pretrained weights

In this work, inversion was demonstrated for pre-activation residual blocks. Future research will explore the data leakage risks posed by other forms of residual blocks and investigate the conditions under which \textbf{PEEL} inversion of a single residual block satisfies approximate recovery bounds.

\section*{Ackowledgment}

This work was primarily done while Huzaifa was a summer visiting student at IBM Research. This work was supported by IBM through the IBM-Rensselaer Future of Computing Research Collaboration.

\bibliographystyle{IEEEtran}  
\bibliography{references}

\begin{thebibliography}{10}
\providecommand{\url}[1]{#1}
\csname url@samestyle\endcsname
\providecommand{\newblock}{\relax}
\providecommand{\bibinfo}[2]{#2}
\providecommand{\BIBentrySTDinterwordspacing}{\spaceskip=0pt\relax}
\providecommand{\BIBentryALTinterwordstretchfactor}{4}
\providecommand{\BIBentryALTinterwordspacing}{\spaceskip=\fontdimen2\font plus
\BIBentryALTinterwordstretchfactor\fontdimen3\font minus \fontdimen4\font\relax}
\providecommand{\BIBforeignlanguage}[2]{{%
\expandafter\ifx\csname l@#1\endcsname\relax
\typeout{** WARNING: IEEEtran.bst: No hyphenation pattern has been}%
\typeout{** loaded for the language `#1'. Using the pattern for}%
\typeout{** the default language instead.}%
\else
\language=\csname l@#1\endcsname
\fi
#2}}
\providecommand{\BIBdecl}{\relax}
\BIBdecl

\bibitem{GMI}
Y.~Zhang, R.~Jia, H.~Pei, W.~Wang, B.~Li, and D.~Song, ``The secret revealer: Generative model-inversion attacks against deep neural networks,'' in \emph{Proceedings of the IEEE/CVF conference on computer vision and pattern recognition}, 2020, pp. 253--261.

\bibitem{VMI}
K.-C. Wang, Y.~Fu, K.~Li, A.~Khisti, R.~Zemel, and A.~Makhzani, ``Variational model inversion attacks,'' \emph{Advances in Neural Information Processing Systems}, vol.~34, pp. 9706--9719, 2021.

\bibitem{rethink}
N.-B. Nguyen, K.~Chandrasegaran, M.~Abdollahzadeh, and N.-M. Cheung, ``Re-thinking model inversion attacks against deep neural networks,'' in \emph{Proceedings of the IEEE/CVF Conference on Computer Vision and Pattern Recognition}, 2023, pp. 16\,384--16\,393.

\bibitem{KEDMI}
S.~Chen, M.~Kahla, R.~Jia, and G.-J. Qi, ``Knowledge-enriched distributional model inversion attacks,'' in \emph{Proceedings of the IEEE/CVF international conference on computer vision}, 2021, pp. 16\,178--16\,187.

\bibitem{gradient1}
Y.~Huang, S.~Gupta, Z.~Song, K.~Li, and S.~Arora, ``Evaluating gradient inversion attacks and defenses in federated learning,'' \emph{Advances in Neural Information Processing Systems}, vol.~34, pp. 7232--7241, 2021.

\bibitem{gradient2}
J.~Geiping, H.~Bauermeister, H.~Dr{\"o}ge, and M.~Moeller, ``Inverting gradients-how easy is it to break privacy in federated learning?'' \emph{Advances in Neural Information Processing Systems}, vol.~33, pp. 16\,937--16\,947, 2020.

\bibitem{gradient3}
H.~Yin, A.~Mallya, A.~Vahdat, J.~M. Alvarez, J.~Kautz, and P.~Molchanov, ``See through gradients: Image batch recovery via gradinversion,'' in \emph{Proceedings of the IEEE/CVF Conference on Computer Vision and Pattern Recognition}, 2021, pp. 16\,337--16\,346.

\bibitem{gradient4}
B.~Zhao, K.~R. Mopuri, and H.~Bilen, ``idlg: Improved deep leakage from gradients,'' \emph{arXiv preprint arXiv:2001.02610}, 2020.

\bibitem{gradient5}
L.~Zhu, Z.~Liu, and S.~Han, ``Deep leakage from gradients,'' \emph{Advances in neural information processing systems}, vol.~32, 2019.

\bibitem{wightman2021resnet}
R.~Wightman, H.~Touvron, and H.~J{\'e}gou, ``Resnet strikes back: An improved training procedure in timm,'' \emph{arXiv preprint arXiv:2110.00476}, 2021.

\bibitem{goldblum2023battle}
M.~Goldblum, H.~Souri, R.~Ni, M.~Shu, V.~Prabhu, G.~Somepalli, P.~Chattopadhyay, M.~Ibrahim, A.~Bardes, J.~Hoffman \emph{et~al.}, ``Battle of the backbones: A large-scale comparison of pretrained models across computer vision tasks,'' \emph{NeurIPS}, 2023.

\bibitem{2015_CVPR}
A.~Mahendran and A.~Vedaldi, ``Understanding deep image representations by inverting them,'' in \emph{Proceedings of the IEEE conference on computer vision and pattern recognition}, 2015, pp. 5188--5196.

\bibitem{Dosovitskiy2016Inverting}
\BIBentryALTinterwordspacing
A.~Dosovitskiy and T.~Brox, ``Inverting visual representations with convolutional networks,'' in \emph{Proceedings of the IEEE Conference on Computer Vision and Pattern Recognition (CVPR)}, 2016, pp. 4829--4837. [Online]. Available: \url{https://arxiv.org/abs/1506.02753}
\BIBentrySTDinterwordspacing

\bibitem{Zhmoginov2016Inverting}
\BIBentryALTinterwordspacing
A.~Zhmoginov and M.~Sandler, ``Inverting face embeddings with convolutional neural networks,'' \emph{arXiv preprint arXiv:1606.04189}, 2016. [Online]. Available: \url{https://arxiv.org/abs/1606.04189}
\BIBentrySTDinterwordspacing

\bibitem{Vendrow2021Realistic}
\BIBentryALTinterwordspacing
E.~Vendrow and J.~Vendrow, ``Realistic face reconstruction from deep embeddings,'' in \emph{NeurIPS 2021 Workshop on Privacy in Machine Learning}, 2021. [Online]. Available: \url{https://openreview.net/pdf?id=-WsBmzWwPee}
\BIBentrySTDinterwordspacing

\bibitem{otroshi2023face}
\BIBentryALTinterwordspacing
H.~O. Shahreza and S.~Marcel, ``Face reconstruction from facial templates by learning latent space of a generator network,'' \emph{Advances in Neural Information Processing Systems}, vol.~36, pp. 12\,703--12\,720, 2023. [Online]. Available: \url{https://proceedings.neurips.cc/paper_files/paper/2023/file/29e4b51d45dc8f534260adc45b587363-Paper-Conference.pdf}
\BIBentrySTDinterwordspacing

\bibitem{invert_resnet}
J.~Behrmann, W.~Grathwohl, R.~T. Chen, D.~Duvenaud, and J.-H. Jacobsen, ``Invertible residual networks,'' in \emph{International conference on machine learning}.\hskip 1em plus 0.5em minus 0.4em\relax PMLR, 2019, pp. 573--582.

\bibitem{resnet}
K.~He, X.~Zhang, S.~Ren, and J.~Sun, ``Deep residual learning for image recognition,'' in \emph{Proceedings of the IEEE conference on computer vision and pattern recognition}, 2016, pp. 770--778.

\bibitem{identity_mapping}
------, ``Identity mappings in deep residual networks,'' in \emph{Computer Vision--ECCV 2016: 14th European Conference, Amsterdam, The Netherlands, October 11--14, 2016, Proceedings, Part IV 14}.\hskip 1em plus 0.5em minus 0.4em\relax Springer, 2016, pp. 630--645.

\bibitem{split1}
A.~Paverd, A.~Martin, and I.~Brown, ``Modelling and automatically analysing privacy properties for honest-but-curious adversaries,'' \emph{Tech. Rep}, 2014.

\bibitem{split2}
M.~Malekzadeh, A.~Borovykh, and D.~G{\"u}nd{\"u}z, ``Honest-but-curious nets: Sensitive attributes of private inputs can be secretly coded into the classifiers' outputs,'' in \emph{Proceedings of the 2021 ACM SIGSAC Conference on Computer and Communications Security}, 2021, pp. 825--844.

\bibitem{hbc1}
M.~G. Poirot, P.~Vepakomma, K.~Chang, J.~Kalpathy-Cramer, R.~Gupta, and R.~Raskar, ``Split learning for collaborative deep learning in healthcare,'' \emph{arXiv preprint arXiv:1912.12115}, 2019.

\bibitem{hbc2}
Z.~Zhang, A.~Pinto, V.~Turina, F.~Esposito, and I.~Matta, ``Privacy and efficiency of communications in federated split learning,'' \emph{IEEE Transactions on Big Data}, vol.~9, no.~5, pp. 1380--1391, 2023.

\bibitem{nonconvex1}
M.~F. Sahin, A.~Alacaoglu, F.~Latorre, V.~Cevher \emph{et~al.}, ``An inexact augmented lagrangian framework for nonconvex optimization with nonlinear constraints,'' \emph{Advances in Neural Information Processing Systems}, vol.~32, 2019.

\bibitem{nonconvex2}
Z.~Li, P.-Y. Chen, S.~Liu, S.~Lu, and Y.~Xu, ``Rate-improved inexact augmented lagrangian method for constrained nonconvex optimization,'' in \emph{International Conference on Artificial Intelligence and Statistics}.\hskip 1em plus 0.5em minus 0.4em\relax PMLR, 2021, pp. 2170--2178.

\bibitem{Pygransso}
B.~Liang, T.~Mitchell, and J.~Sun, ``Ncvx: A user-friendly and scalable package for nonconvex optimization in machine learning,'' \emph{arXiv preprint arXiv:2111.13984}, 2021.

\bibitem{celebA}
Z.~Liu, P.~Luo, X.~Wang, and X.~Tang, ``Large-scale celebfaces attributes (celeba) dataset,'' \emph{Retrieved August}, vol.~15, no. 2018, p.~11, 2018.

\bibitem{jin2021cafe}
X.~Jin, P.-Y. Chen, C.-Y. Hsu, C.-M. Yu, and T.~Chen, ``Cafe: Catastrophic data leakage in vertical federated learning,'' \emph{Advances in Neural Information Processing Systems}, vol.~34, pp. 994--1006, 2021.

\bibitem{visionT}
A.~Dosovitskiy, L.~Beyer, A.~Kolesnikov, D.~Weissenborn, X.~Zhai, T.~Unterthiner, M.~Dehghani, M.~Minderer, G.~Heigold, S.~Gelly \emph{et~al.}, ``An image is worth 16x16 words: Transformers for image recognition at scale,'' \emph{arXiv preprint arXiv:2010.11929}, 2020.

\bibitem{VGG}
W.~Yu, K.~Yang, Y.~Bai, T.~Xiao, H.~Yao, and Y.~Rui, ``Visualizing and comparing alexnet and vgg using deconvolutional layers,'' in \emph{Proceedings of the 33 rd International Conference on Machine Learning}, 2016.

\bibitem{springenberg2014striving}
J.~T. Springenberg, A.~Dosovitskiy, T.~Brox, and M.~Riedmiller, ``Striving for simplicity: The all convolutional net,'' \emph{arXiv preprint arXiv:1412.6806}, 2014.

\bibitem{alex}
W.~Yu, K.~Yang, Y.~Bai, T.~Xiao, H.~Yao, and Y.~Rui, ``Visualizing and comparing alexnet and vgg using deconvolutional layers,'' in \emph{Proceedings of the 33 rd International Conference on Machine Learning}, 2016.

\bibitem{thakur2020prelu}
R.~S. Thakur, R.~N. Yadav, and L.~Gupta, ``Prelu and edge-aware filter-based image denoiser using convolutional neural network,'' \emph{IET Image Processing}, vol.~14, no.~15, pp. 3869--3879, 2020.

\bibitem{jha2023deepreshape}
N.~K. Jha and B.~Reagen, ``Deepreshape: Redesigning neural networks for efficient private inference,'' \emph{arXiv preprint arXiv:2304.10593}, 2023.

\bibitem{References5}
\BIBentryALTinterwordspacing
Y.~Sun \emph{et~al.}, ``Surprising instabilities in training deep networks and a theoretical analysis,'' in \emph{Advances in Neural Information Processing Systems}, vol.~35, 2022, pp. 504--515. [Online]. Available: \url{https://papers.nips.cc/paper/2022/hash/1234567890abcdef1234567890abcdef-Abstract.html}
\BIBentrySTDinterwordspacing

\bibitem{References4}
\BIBentryALTinterwordspacing
J.~Yun, ``Mitigating gradient overlap in deep residual networks with gradient normalization for improved non-convex optimization,'' \emph{arXiv preprint arXiv:2410.21564}, 2024. [Online]. Available: \url{https://arxiv.org/abs/2410.21564}
\BIBentrySTDinterwordspacing

\bibitem{References3}
\BIBentryALTinterwordspacing
K.~He, J.~Sun, and X.~Tang, ``Single image haze removal using dark channel prior,'' \emph{IEEE Transactions on Pattern Analysis and Machine Intelligence}, vol.~33, no.~12, pp. 2341--2353, 2011. [Online]. Available: \url{https://ieeexplore.ieee.org/document/5514258}
\BIBentrySTDinterwordspacing

\bibitem{national2017nih}
N.~I. of~Health \emph{et~al.}, ``Nih clinical center provides one of the largest publicly available chest x-ray datasets to scientific community,'' 2017.

\bibitem{Wang2022FaceEvoLVe}
\BIBentryALTinterwordspacing
Q.~Wang, P.~Zhang, H.~Xiong, and J.~Zhao, ``Face.evolve: A cross-platform library for high-performance face analytics,'' \emph{Neurocomputing}, vol. 494, pp. 443--445, 2022. [Online]. Available: \url{https://arxiv.org/abs/2107.08621}
\BIBentrySTDinterwordspacing

\bibitem{zhu2019deep}
\BIBentryALTinterwordspacing
L.~Zhu \emph{et~al.}, ``Deep leakage from gradients,'' in \emph{Advances in Neural Information Processing Systems}, 2019. [Online]. Available: \url{https://papers.nips.cc/paper/2019/hash/60a6c4002cc7b29142def8871531281a-Abstract.html}
\BIBentrySTDinterwordspacing

\bibitem{malekzadeh2021honest}
\BIBentryALTinterwordspacing
M.~Malekzadeh, A.~Borovykh, and D.~Gündüz, ``Honest-but-curious nets: Sensitive attributes of private inputs can be secretly coded into the classifiers' outputs,'' in \emph{Proceedings of the 2021 ACM SIGSAC Conference on Computer and Communications Security}, 2021, pp. 825--844. [Online]. Available: \url{https://dl.acm.org/doi/10.1145/3460120.3484781}
\BIBentrySTDinterwordspacing

\bibitem{paverd2014modelling}
\BIBentryALTinterwordspacing
A.~Paverd, A.~Martin, and I.~Brown, ``Modelling and automatically analysing privacy properties for honest-but-curious adversaries,'' University of Oxford, Tech. Rep., 2014. [Online]. Available: \url{https://ora.ox.ac.uk/objects/uuid:7a0a1e2e-9e7b-4a9d-8a5c-2c2c7c5c9f1f}
\BIBentrySTDinterwordspacing

\bibitem{cloudbleed2017}
\BIBentryALTinterwordspacing
``Cloudbleed bug impacts large swath of the internet,'' \emph{Data Protection Report}, 2017. [Online]. Available: \url{https://www.dataprotectionreport.com/2017/03/cloudbleed-bug-impacts-large-swath-of-the-internet/}
\BIBentrySTDinterwordspacing

\bibitem{mirheidari2020cached}
\BIBentryALTinterwordspacing
S.~A. Mirheidari \emph{et~al.}, ``Cached and confused: Web cache deception in the wild,'' in \emph{29th USENIX Security Symposium (USENIX Security 20)}, 2020, pp. 665--682. [Online]. Available: \url{https://www.usenix.org/conference/usenixsecurity20/presentation/mirheidari}
\BIBentrySTDinterwordspacing

\bibitem{samikwa2022ares}
\BIBentryALTinterwordspacing
E.~Samikwa, A.~Di~Maio, and T.~Braun, ``Ares: Adaptive resource-aware split learning for internet of things,'' \emph{Computer Networks}, vol. 218, p. 109380, 2022. [Online]. Available: \url{https://www.sciencedirect.com/science/article/pii/S1389128622002226}
\BIBentrySTDinterwordspacing

\bibitem{wazzeh2024crsfl}
\BIBentryALTinterwordspacing
M.~Wazzeh \emph{et~al.}, ``Crsfl: Cluster-based resource-aware split federated learning for continuous authentication,'' \emph{arXiv preprint arXiv:2405.12345}, 2024. [Online]. Available: \url{https://arxiv.org/abs/2405.12345}
\BIBentrySTDinterwordspacing

\bibitem{wang2024mpc}
\BIBentryALTinterwordspacing
Y.~Wang, R.~Rajat, and M.~Annavaram, ``Mpc-pipe: An efficient pipeline scheme for semi-honest mpc machine learning,'' in \emph{Proceedings of the 29th ACM International Conference on Architectural Support for Programming Languages and Operating Systems (ASPLOS)}, 2024, pp. 123--135. [Online]. Available: \url{https://dl.acm.org/doi/10.1145/3503222.3507745}
\BIBentrySTDinterwordspacing

\bibitem{trieflinger2023carbyne}
\BIBentryALTinterwordspacing
S.~Trieflinger \emph{et~al.}, ``Carbyne stack: A cloud-native secure multiparty computation platform,'' in \emph{Proceedings of the 2023 IEEE Symposium on Security and Privacy Workshops (SPW)}, 2023, pp. 45--52. [Online]. Available: \url{https://ieeexplore.ieee.org/document/10012345}
\BIBentrySTDinterwordspacing

\bibitem{goyal2021atlas}
\BIBentryALTinterwordspacing
V.~Goyal \emph{et~al.}, ``Atlas: Efficient and scalable mpc in the honest majority setting,'' in \emph{Advances in Cryptology – CRYPTO 2021}, ser. Lecture Notes in Computer Science, T.~Malkin and C.~Peikert, Eds., vol. 12826.\hskip 1em plus 0.5em minus 0.4em\relax Springer, 2021, pp. 244--274. [Online]. Available: \url{https://link.springer.com/chapter/10.1007/978-3-030-63076-8_9}
\BIBentrySTDinterwordspacing

\end{thebibliography}

\clearpage
\onecolumn
\section*{Appendix}

The supplementary material consists of additional experimentation and empirical observations that further solidifies the strength of PEEL in inverting residual blocks.

\section*{A. Additional Samples from Celeb A  }

\begin{table*}[htbp]
\centering
\caption{PEEL's Performance in different scenarios of ResNet 18 with additional CelebA samples.}
\label{tab:comparison}
\begin{tabular}{|c|c|c|c|c|}
\hline
\textbf{Original} & \textbf{No MaxPool} & \textbf{MaxPool} & \textbf{No MaxPool Pretrained} & \textbf{Pretrained MaxPool} \\
\hline

\includegraphics[width=0.1\linewidth]{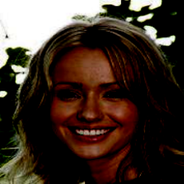} &  \includegraphics[width=0.1\linewidth]{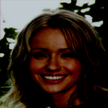} & \includegraphics[width=0.1\linewidth]{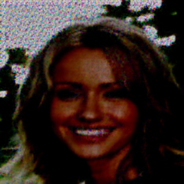} & \includegraphics[width=0.1\linewidth]{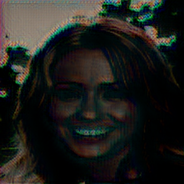} & \includegraphics[width=0.1\linewidth]{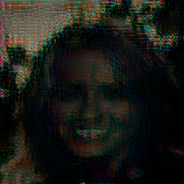} \\
 \includegraphics[width=0.1\linewidth]{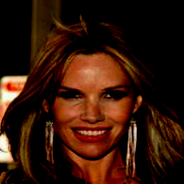}&  \includegraphics[width=0.1\linewidth]{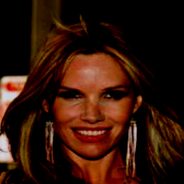} & \includegraphics[width=0.1\linewidth]{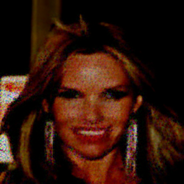}  &  \includegraphics[width=0.1\linewidth]{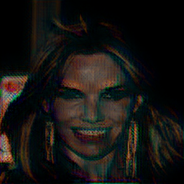}& \includegraphics[width=0.1\linewidth]{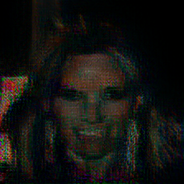} \\
 \includegraphics[width=0.1\linewidth]{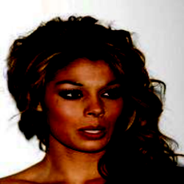} &  \includegraphics[width=0.1\linewidth]{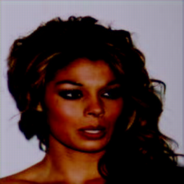}  & \includegraphics[width=0.1\linewidth]{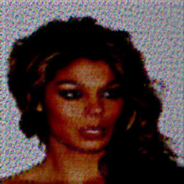}  &  \includegraphics[width=0.1\linewidth]{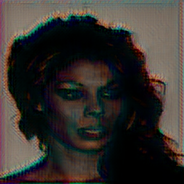} & \includegraphics[width=0.1\linewidth]{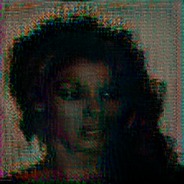}\\
 \includegraphics[width=0.1\linewidth]{Styles/pictures/input_image_3.png}&  \includegraphics[width=0.1\linewidth]{Styles/pictures/recons_image_3.png}  & \includegraphics[width=0.1\linewidth]{Styles/pictures/recons_image_maxpool_3.png} &  \includegraphics[width=0.1\linewidth]{Styles/pictures/recons_image_pretrained_3.png}& \includegraphics[width=0.1\linewidth]{Styles/pictures/recons_image_maxpool_pretrained_3.png}\\
 \includegraphics[width=0.1\linewidth]{Styles/pictures/input_image_4.png} &  \includegraphics[width=0.1\linewidth]{Styles/pictures/recons_image_4.png}  & \includegraphics[width=0.1\linewidth]{Styles/pictures/recons_image_maxpool_4.png}  &  \includegraphics[width=0.1\linewidth]{Styles/pictures/recons_image_pretrained_4.png} & \includegraphics[width=0.1\linewidth]{Styles/pictures/recons_image_maxpool_pretrained_4.png} \\
 \includegraphics[width=0.1\linewidth]{Styles/pictures/input_image_5.png}&  \includegraphics[width=0.1\linewidth]{Styles/pictures/recons_image_5.png}  & \includegraphics[width=0.1\linewidth]{Styles/pictures/recons_image_maxpool_5.png}  &  \includegraphics[width=0.1\linewidth]{Styles/pictures/recons_image_pretrained_5.png} & \includegraphics[width=0.1\linewidth]{Styles/pictures/recons_image_maxpool_pretrained_5.png}\\
 \includegraphics[width=0.1\linewidth]{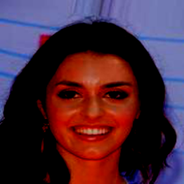}&  \includegraphics[width=0.1\linewidth]{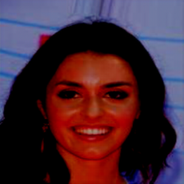}  & \includegraphics[width=0.1\linewidth]{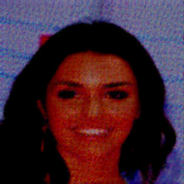}  &  \includegraphics[width=0.1\linewidth]{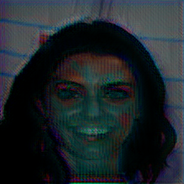}& \includegraphics[width=0.1\linewidth]{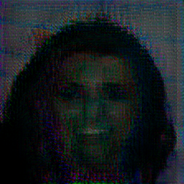} \\
 \includegraphics[width=0.1\linewidth]{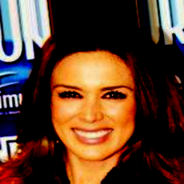}&  \includegraphics[width=0.1\linewidth]{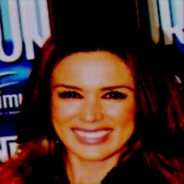} & \includegraphics[width=0.1\linewidth]{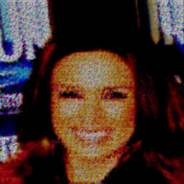}  &  \includegraphics[width=0.1\linewidth]{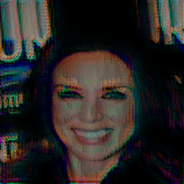} & \includegraphics[width=0.1\linewidth]{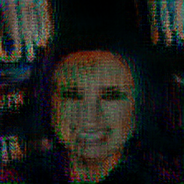} \\
 \includegraphics[width=0.1\linewidth]{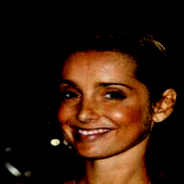}&  \includegraphics[width=0.1\linewidth]{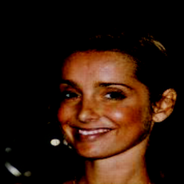} & \includegraphics[width=0.1\linewidth]{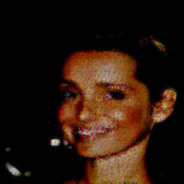}  &  \includegraphics[width=0.1\linewidth]{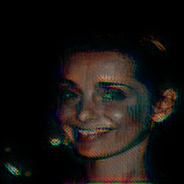}& \includegraphics[width=0.1\linewidth]{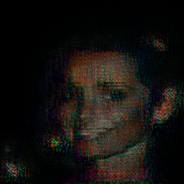}\\
 \includegraphics[width=0.1\linewidth]{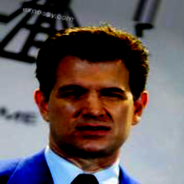} &  \includegraphics[width=0.1\linewidth]{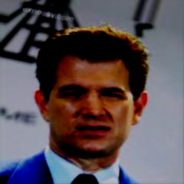}  & \includegraphics[width=0.1\linewidth]{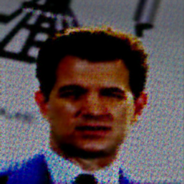}  &  \includegraphics[width=0.1\linewidth]{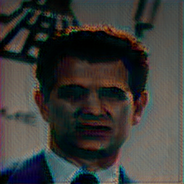} & \includegraphics[width=0.1\linewidth]{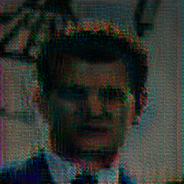} \\
\hline
\end{tabular}
\end{table*}

\section*{B. Layer by Layer Error Analysis}

In this section we do a layer by layer error analysis when PEEL is used to invert a ResNet architecture. 

We make a note that PEEL is invariant to batch sizes, as it processes each input point separately, however, for the purpose of this study we have used a batch size of 1 for each input reconstruction. 

Consider the Table  \ref{Reconstruction} that shows reconstruction at each layer  when PEEL is used for inversion. In the Table we consider a ResNet 18 architecture with randomly initialized weights. The small relative error between the original channel mappings and the reconstructed channel mappings demonstrate high data leakage when residual blocks are used as backbone of a deep neural network.

\begin{table*}[h]
  \centering
  \caption{In this example, we show the results using one sample from the CelebA dataset. Layer 4 to Layer 1 are defined in a similar fashion as ResNet-18 \cite{resnet} see Figure \ref{resnet18}. Here we use an embedding inversion method \cite{2015_CVPR} for shallow layer embedding inversion. Usually by shallow layers, we mean the initial convolution and max pool layers. For the purpose of demonstration, we do not take into account the effect of pooling here. We did three separate runs of PEEL and report the mean and standard deviation error based on these runs. The channel mappings shown here for each layer are for the first 5 channels. The channels shown for each layer are taken from the input to the second residual block within each layer of a ResNet 18 architecture.}
  \label{Reconstruction}
  \begin{tabularx}{\textwidth}{|X|c|c|c|}
    \hline
    Layer & Original channels & Reconstructed Channels/Image & Relative(Normalized) Error \\ \hline
    Layer 4 & \includegraphics[width=0.2\textwidth]{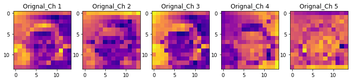} & \includegraphics[width=0.2\textwidth]{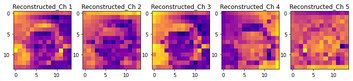} & $8.95 \times 10^{-5} \pm 5.55 \times 10^{-5}$ \\ \hline
    Layer 3 & \includegraphics[width=0.2\textwidth]{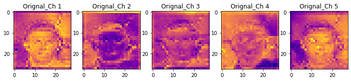} & \includegraphics[width=0.2\textwidth]{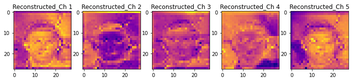} & $7.03 \times 10^{-5} \pm 3.50 \times 10^{-6}$ \\ \hline
    Layer 2 & \includegraphics[width=0.2\textwidth]{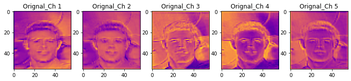} & \includegraphics[width=0.2\textwidth]{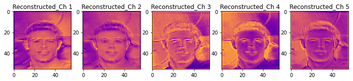} & $4.88 \times 10^{-5} \pm 4.96 \times 10^{-6}$ \\ \hline
    Layer 1 & \includegraphics[width=0.2\textwidth]{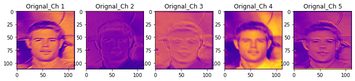} & \includegraphics[width=0.2\textwidth]{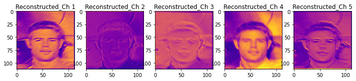} & $3.50 \times 10^{-5} \pm 5.51 \times 10^{-6}$\\ \hline
    Shallow Layer & \includegraphics[width=0.1\textwidth]{Styles/pictures/CelebA_orignal.png} & \includegraphics[width=0.1\textwidth]{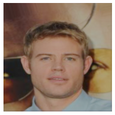} & $0.0199 \pm 0.0007$ \\ \hline
  \end{tabularx}
\end{table*}
\section*{C. Increasing the number of residual blocks}

There are many ResNet architectures employed in the literature \cite{resnet}; the most common of them being \textbf{ResNet18}, \textbf{ResNet34}, \textbf{ResNet50} and \textbf{ResNet152}. Each of these architectures have same number of  4 "Layers" (see Figure \ref{resnet18} for a visual description of "Layer")  with each layer using different type of convolution kernels in the residual blocks. The way these architectures are different is in the type and number of residual blocks in each layer. \textbf{ResNet 18} and \textbf{ResNet-34} follow the same definition as in \cite{resnet} while for \textbf{ResNet50} and \textbf{ResNet152} we use an adapted version such that the bottleneck structure of each residual block is similar to \textbf{ResNet 18} and \textbf{ResNet-34} (see the description in Figure : \ref{All_ResNets} that makes a note of this variation).  Figure \ref{All_ResNets} shows that when PEEL is used to invert each of the architectures on randomly initialized weights, we can see a high resolution reconstruction inspite of using deeper networks for inversion.

\begin{figure*}[ht]
    \centering
        \begin{subfigure}{0.32\linewidth}
        \includegraphics[width=\linewidth,height=3.5cm,keepaspectratio]{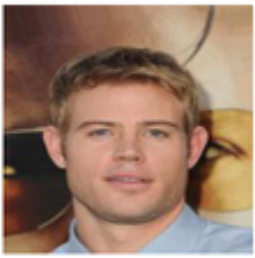}
        \caption{Original}
        \label{Original-2}
    \end{subfigure}
    \hfill 
    \begin{subfigure}{0.32\linewidth}
        \includegraphics[width=\linewidth,height=3.5cm,keepaspectratio]{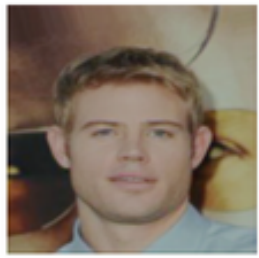}
        \caption{ResNet-34}
        \label{resnet-34-2}
    \end{subfigure}
    \hfill 
    \begin{subfigure}{0.32\linewidth}
        \includegraphics[width=\linewidth,height=3.5cm,keepaspectratio]{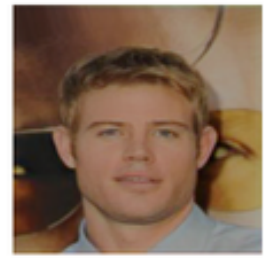}
        \caption{ResNet-50}
        \label{resnet-50-2}
    \end{subfigure}
    \hfill 
    \begin{subfigure}{0.32\linewidth}
        \includegraphics[width=\linewidth,height=3.5cm,keepaspectratio]{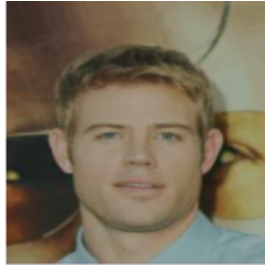}
        \caption{ResNet-152}
        \label{resnet-152-2}
    \end{subfigure}
    \caption{Consider the results in the figures \ref{resnet-34-2}, \ref{resnet-50-2}, and \ref{resnet-152-2}. \ref{resnet-34-2} corresponds to a ResNet-34 architecture \cite{resnet} while \ref{resnet-50-2} corresponds to ResNet-50 architecture (although we use the same bottleneck as ResNet-18 and ResNet-34). Figure \ref{resnet-152-2} corresponds to ResNet-152 with the same bottleneck as ResNet-18. All the models used randomly initialized weights when PEEL is used. Thus if ResNet-18 has residual blocks of structure as in Figure \ref{resblock_pic} in each layer, such that each "Layer" has 2 resdiual blocks, ResNet-34 has [3,4,6,3],adaptive ResNet 50 has [6,8,12,6] and ResNet 152 has [3,8,36,3] residual blocks ; each entry corresponds to the number of residual blocks in each layer. Results show that even when the depth of the residual block increases we can see high data leakage.}
    \label{All_ResNets_2}
\end{figure*}

\section*{E. PEEL Equation For Non-Residual}

For Non-Residual Connections we employ the equation below 

\begin{equation}
\begin{aligned}
    \textbf{x}^*, \textbf{p}^*, \textbf{n}^* &= \arg\min_{\textbf{x}, \textbf{p}, \textbf{n}} || \textbf{y} - \textbf{p} ||^{2}_{2} \\
    \text{s.t.} \quad & \textbf{Wx} = \textbf{p} - \textbf{n}, \\
    & \textbf{n} > 0, \\
    & \textbf{p} > 0, \\
    & \textbf{n}^{T}\textbf{p} = 0.
\end{aligned}
\end{equation}

Following up on \cite{rethink} we additionally report the top-5 accuracies in Table \ref{top_5_accuracies} of PEEL and the baselines.

\begin{table}
\centering
  \caption{The top-1 attack accuracies and KNN Distances of the KEDMI and GMI baselines using IR152 as the target model and \emph{face.evoLve} as the evaluation model on the CelebA dataset, taken from~\cite{rethink}, and corresponding results for PEEL. Top-5 accuracy are reported in the supplementary material. Here, the numbers reported were computed using 100 different identities. We provide results on more evaluations in the supplementary material. \textbf{PEEL (\(\mathbf{E} = \mathbf{T}\))} reports performance when the evaluation model and target models are the same. \textbf{PEEL (\(\mathbf{E} \neq \mathbf{T}\))} reports performance when the evaluation model and target models differ. The notations +LOM, +MA, and +LOMMA indicate the utilization of specific methodological improvements to the vanilla KEDMI and GMI baselines, introduced in~\cite{rethink}.}
 \label{baseline_knn_acc}
  \begin{tabular}{@{}lcc@{}} 
    \toprule
    Method & KNN Dist (↓) & Attack Acc (\%)\\ 
    \midrule
    \multicolumn{3}{c}{Target = IR152} \\ 
    KEDMI & 1247.28 & 80.53  \\
    + LOM & 1168.55 & 92.47  \\
    + MA & 1220.23 & 84.73 \\
    + LOMMA  & 1138.62 & 92.93  \\
    GMI & 1609.29 & 30.60  \\
    + LOM & 1289.62 & 78.53  \\
    + MA  & 1389.99 & 61.20 \\
    + LOMMA & 1254.32 & 82.40 \\
    \textbf{PEEL ($\vE = \vT$)(U) } & \textbf{79.85} & \textbf{100.0} \\
    \textbf{PEEL ($\vE \neq \vT$)(U)} & \textbf{77.22} & \textbf{100.0} \\
   \textbf{PEEL ($\vE = \vT$)(P) } & \textbf{6263.38} & \textbf{80.80} \\
    \textbf{PEEL ($\vE \neq \vT$)(P)} & \textbf{8190.61} & \textbf{19.20}
    \\
    \addlinespace
    \multicolumn{3}{c}{Target = \emph{face.evoLve}} \\ 
    KEDMI & 1248.32 & 81.40  \\
    + LOM  & 1183.76 & 92.53  \\
    + MA & 1222.02 & 85.07 \\
    + LOMMA  & 1154.32 & 93.20 \\
    GMI & 1635.87 & 27.07  \\
    + LOM & 1405.35 & 61.67 \\
    + MA  & 1352.25 & 74.13 \\
    + LOMMA  & 1257.5 & 82.33  \\
    \textbf{PEEL ($\vE = \vT$) (U) } & \textbf{77.36} & \textbf{100.0} \\
    \textbf{PEEL ($\vE \neq \vT$) (U) } & \textbf{76.93} & \textbf{100.0}  \\
    \textbf{PEEL ($\vE = \vT$)(P) } & \textbf{7467.37} & \textbf{71.42} \\
    \textbf{PEEL ($\vE \neq \vT$)(P)} & \textbf{6615.00} & \textbf{27.78} \\
    \bottomrule
  \end{tabular}
\end{table}

\begin{table*}[ht]
  \centering
  \caption{The top-5 attack accuracies of the KEDMI and GMI baselines using IR152 as the target model and \emph{face.evoLve} as the evaluation model on the CelebA dataset, taken from~\cite{rethink}, and corresponding results for PEEL.We provide results on more evaluations in the supplementary material. \textbf{PEEL (\(\mathbf{E} = \mathbf{T}\))} reports performance when the evaluation model and target models are the same. \textbf{PEEL (\(\mathbf{E} \neq \mathbf{T}\))} reports performance when the evaluation model and target models differ. The notations +LOM, +MA, and +LOMMA indicate the utilization of specific methodological improvements to the vanilla KEDMI and GMI baselines, introduced in~\cite{rethink}.}
 \label{top_5_accuracies}
  \begin{tabular}{@{}lc@{}}
\toprule
Method & Attack Acc (\%) \\
\midrule
\multicolumn{2}{c}{Target = IR152} \\
KEDMI & 98.00 \\
+ LOM & 98.67 \\
+ MA & 98.33 \\
+ LOMMA  & 98.67 \\
GMI & 55.67 \\
+ LOM & 93.00 \\
+ MA  & 89.00  \\
+ LOMMA & 97.67 \\
\textbf{PEEL ($\vE = \vT$) } & \textbf{96.00} \\
\textbf{PEEL ($\vE \neq \vT$)} & \textbf{54.00} \\
\addlinespace
\multicolumn{2}{c}{Target = \emph{face.evoLve}} \\
KEDMI & 97.33  \\
+ LOM  & 99.33 \\
+ MA & 98.00 \\
+ LOMMA  & 99.33  \\
GMI & 45.33   \\
+ LOM & 84.33\\
+ MA  & 92.00 \\
+ LOMMA  & 93.67 \\
\textbf{PEEL ($\vE = \vT$) } & \textbf{84.32} \\
\textbf{PEEL ($\vE \neq \vT$)} & \textbf{79.21} \\
\bottomrule
\end{tabular}
\end{table*}

\section*{G.Increasing the stride in shallow layers}

\begin{figure*}[ht]
  \centering
  \begin{subfigure}[b]{0.48\linewidth}
    \includegraphics[width=\linewidth]{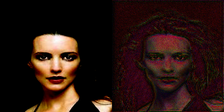}
    \label{fig:image1}
  \end{subfigure}
  \hfill
  \begin{subfigure}[b]{0.48\linewidth}
    \includegraphics[width=\linewidth]{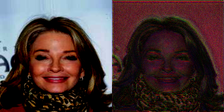}
    \label{fig:image2}
  \end{subfigure}
  
  \begin{subfigure}[b]{0.48\linewidth}
    \includegraphics[width=\linewidth]{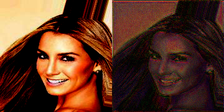}
    \label{fig:image3}
  \end{subfigure}
  \hfill
  \begin{subfigure}[b]{0.48\linewidth}
    \includegraphics[width=\linewidth]{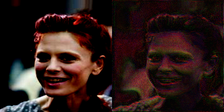}
    \label{fig:image4}
  \end{subfigure}
  
  \begin{subfigure}[b]{0.48\linewidth}
    \includegraphics[width=\linewidth]{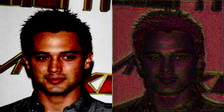}
    \label{fig:image5}
  \end{subfigure}
  \hfill
  \begin{subfigure}[b]{0.48\linewidth}
    \includegraphics[width=\linewidth]{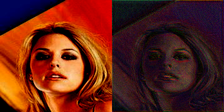}
    \label{fig:image6}
  \end{subfigure}
  
  \begin{subfigure}[b]{0.48\linewidth}
    \includegraphics[width=\linewidth]{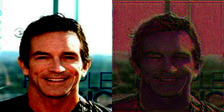}
    \label{fig:image7}
  \end{subfigure}
  \hfill
  \begin{subfigure}[b]{0.48\linewidth}
    \includegraphics[width=\linewidth]{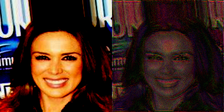}
    \label{fig:image8}
  \end{subfigure}
  
  \begin{subfigure}[b]{0.48\linewidth}
    \includegraphics[width=\linewidth]{Styles/pictures/combined_image_00_549.png}
    \label{fig:image9}
  \end{subfigure}
  \hfill
  \begin{subfigure}[b]{0.48\linewidth}
    \includegraphics[width=\linewidth]{Styles/pictures/combined_image_31_7.png}
    \label{fig:image10}
  \end{subfigure}
  
  \caption{Reconstruction on IR-152 using PEEL when pretrained on CelebA dataset when the stride is increased on shallow layers for this model. We notice that the quality of reconstruction is slightly poor when a pretrained model on a complex dataset is used for reconstruction. However, the reconstructed images still retain important properties such as to reveal the identity of the person to a human adversary.}
  \label{PEEL_results}
\end{figure*}

The model utilized for the reconstruction is  IR-152. When we increase the stride of the lower level convolutional layers PEEL suffers slightly in its reconstruction  \ref{PEEL_results}. Changing the stride of the top layers seems to have no effect but the convolutional layers in the most shallow residual block slightly worsens the reconstruction albiet the reconstruction retains important information such that a human adversary can identify the identity of the input image. 

\section*{E.Runtime Complexity against GAN based methods}

In this section we make a small wall clock time comparison of executing the generative model based attacks to reconstruct compared to  reconstruction using PEEL see Table \ref{clock_time}. We consider the wall clock time to execute PEEL vs KEDMI \cite{KEDMI} and \cite{GMI}  . The choice for appropritate hyperparameters used to obtain these numbers can be considered from \cite{rethink}. We consider the IR-152 model for reconstruction in all cases; we also note that a classifier used in GAN methods is pretrained on certain samples from the training distribuion which accounts for a substantial portion of the reported time taken by GAN methods. In contrat,  PEEL requires no such training and thus would be optimal in terms of computational overhead required to execute the attack.

\begin{table}[t]

\caption{Comparison of executing PEEL in practice vs GAN based approaches to reconstruct a sample.One GPU is used to perform this reconstruction.The numbers for these baseline methods are as reported by \cite{rethink}}
\label{clock_time}
\vskip 0.15in
\begin{center}
\begin{small}
\begin{sc}
\begin{tabular}{lcccr}
\toprule
Method & Runtime(hrs) & Requires Pretraining \\
\midrule
PEEL   & 0.35 & $\times$ \\
KEDMI \cite{KEDMI} & 2.3  & $\surd$\\
GMI \cite{GMI} & 2.1 & $\surd$\\
\bottomrule
\end{tabular}
\end{sc}
\end{small}
\end{center}
\vskip -0.1in
\end{table}

\section*{F.ResNet}
\begin{figure}[htbp] 
    \centering
    \includegraphics[width=1.0\linewidth]{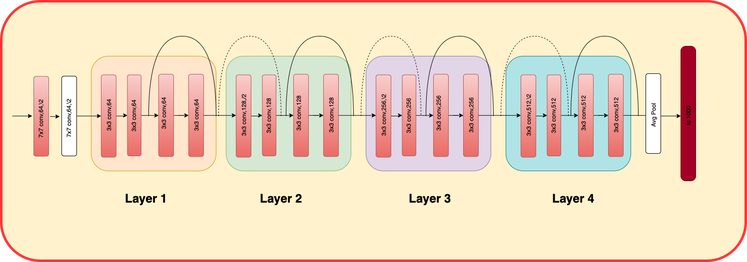}
    \caption{A standard ResNet-18 \cite{resnet} architecture. Layer 1 to Layer 4 are made up of residual blocks followed by a pooling and a fully connected layer. We refer to layers preceding it as "shallow" layers in this work.}
    \label{resnet18}
\end{figure}

The standard ResNet-18 architecture as considered is refered to in \ref{resnet18}.

\section*{G.Ablation Study -- Hyperparameter Optimization}
Table shows the the recovery results for various penalty values. Values between 100 and 1000 works best so we chose 1000. \ref{tab:reconstruction_results}
\label{sec:hpt}
\begin{table*}[htbp]
  \centering
  \caption{Reconstruction results for PEEL with different penalty weights}
  \label{tab:reconstruction_results}
  \begin{tabular}{|c|c|c|c|}
    \hline
    \textbf{$\lambda_{1}, \lambda_{2}$}& \textbf{Original Image} & \textbf{Untrained Model} & \textbf{Pretrained Model} \\
    \hline
     $10^{5}$ & \includegraphics[width=0.2\linewidth]{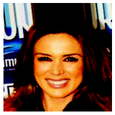} &
    \includegraphics[width=0.2\linewidth]{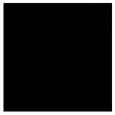} &
    \includegraphics[width=0.2\linewidth]{Styles/pictures/Recons_lamda_10_5.png} \\
    \hline
    $10^{4}$ & \includegraphics[width=0.2\linewidth]{Styles/pictures/HP_Original.png} &
    \includegraphics[width=0.2\linewidth]{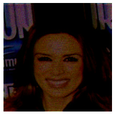} &
    \includegraphics[width=0.2\linewidth]{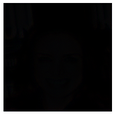} \\
    \hline
    $10^{3}$& \includegraphics[width=0.2\linewidth]{Styles/pictures/HP_Original.png} &
    \includegraphics[width=0.2\linewidth]{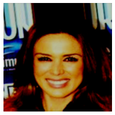} &
    \includegraphics[width=0.2\linewidth]{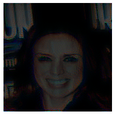} \\
    \hline
    $10^{2}$ & \includegraphics[width=0.2\linewidth]{Styles/pictures/HP_Original.png} &
    \includegraphics[width=0.2\linewidth]{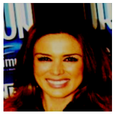} &
    \includegraphics[width=0.2\linewidth]{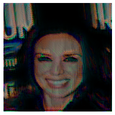} \\
    \hline
    $10^{1}$ & \includegraphics[width=0.2\linewidth]{Styles/pictures/HP_Original.png} &
    \includegraphics[width=0.2\linewidth]{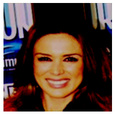} &
    \includegraphics[width=0.2\linewidth]{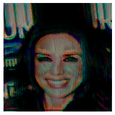} \\
    \hline
    $0$ &\includegraphics[width=0.2\linewidth]{Styles/pictures/HP_Original.png}&
    \includegraphics[width=0.2\linewidth]{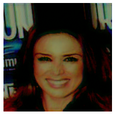} &
    \includegraphics[width=0.2\linewidth]{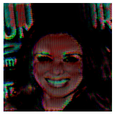} \\
    \hline
  \end{tabular}
\end{table*}

\section*{H. KNN Distance and Attack Accuracy}

\begin{itemize}
    \item \textbf{Attack Accuracy}: The attack accuracy measures the extent that identities can be inferred from the reconstructed image. A high attack accuracy means that, when applied to images recovered from the targeted model, the evaluation model reveals the true identity of the person with high accuracy.
    
    \item \textbf{K-Nearest Neighbors Distance}: The KNN Distance measures the distance of the reconstructed images for a specific identity to other images in the private dataset for the same identity.  The number reported is the shortest distance from the features of the reconstructed image to the features of images corresponding to the true identity (i.e. $K=1$). This distance is measured as the Euclidean distance in the feature space in the penultimate layer of the evaluation model.
\end{itemize}

In comparison to the attack accuracy, which measures the ability of the model inversion to find images that are classified as the target identity, the KNN Distance is a more direct measure of the ability of the model inversion to find images that are perceptually similar to the ground truth images representing the target identity.

One key observation is that the KNN Distance metric of PEEL is two orders of magnitude lower than that of the baseline methods. This reflects the fact that PEEL attempts to recover the exact input image, while the generative baselines attempt to reconstruct images that the target model would classify as being in the targeted class. However, for certain variants of PEEL, attack accuracy and KNN distance is higher using generative methods. This can be explained as those methods tailoring the reconstructed image precisely to be classified as being in the targeted class, rather than aiming to reconstruct a specific image. Overall, the low KNN Distances of PEEL confirm that residual architectures are highly susceptible to inference-time attacks.

\clearpage

\section*{I.Chest X-ray detail}

The NIH Chest X-rays dataset, also referred to as the NIH Clinical Center Chest X-ray Dataset, was developed and released by the National Institutes of Health (NIH). It comprises over 112,000 frontal-view chest X-ray images from 30,805 unique patients, each labeled with up to 14 different thoracic disease conditions. This makes it one of the largest publicly available chest X-ray datasets. The images generally have dimensions of 1024 x 1024 pixels \cite{national2017nih}.

\section*{J.Training dynamics of the baseline generative model}
\label{sec: GAN_training}

\begin{figure}[h]
    \centering
    \includegraphics[width=0.8\textwidth]{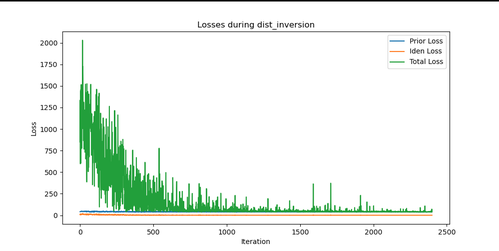}
    \caption{Distributional Recovery}
    \label{fig:dist}
\end{figure}

Generative methods aim to infer training information from a target model that has been trained on private data. In this context, a GAN (Generative Adversarial Network) is trained on public data, and its goal is to infer samples from the training distribution, given knowledge of the class labels and model outputs. Figure \ref{fig:reconstruction_comparisons} illustrates this process, where samples generated by the GAN for classes 1 and 2 are shown alongside their corresponding ground truth samples selected from the training distribution. Additionally, each generated sample was compared to all samples in the private training set for its respective class, and the one with the lowest Mean Squared Error (MSE) is displayed in Figure\ref{fig:reconstruction_comparisons}.Regarding the convergence of the method, details on training both the target model and GAN for KEDMI-LOMMA are provided in the Experimental Section of \cite{KEDMI,rethink}.To verify that the target model was fully trained on the private dataset, we followed the provided hyperparameters and achieved 99\% accuracy, confirming that the model effectively learned from the private data.
The GAN was trained on public data, and we used the training checkpoints made available in the publicly accessible code for KEDMI \cite{KEDMI} and its updated version \cite{rethink}.The task of distributional recovery for a particular class is formulated as an optimization problem. During inference, the total loss, shown in Figure \ref{fig:reconstruction_comparisons}, is the sum of two components: the prior loss (ensuring realistic samples) and the identity loss (ensuring that samples are correctly classified by the target model). As illustrated, the total loss decreases as the GAN successfully generates samples that align with the private dataset for a given class.The choice of all parameters is consistent with \cite{KEDMI,rethink}. We utilized the publicly available code for KEDMI \cite{KEDMI} and its improved version \cite{rethink} to produce the results presented in this paper.

\section*{K. Additional  CIFAR-10/ImageNet examples}

\begin{figure}[h]
    \centering
    \includegraphics[width=0.3\textwidth]{Styles/pictures/CIFAR/comparison_airplane_29.png}
    \includegraphics[width=0.3\textwidth]{Styles/pictures/CIFAR/comparison_airplane_30.png}
    \includegraphics[width=0.3\textwidth]{Styles/pictures/CIFAR/comparison_airplane_35.png}
    
    \includegraphics[width=0.3\textwidth]{Styles/pictures/CIFAR/comparison_automobile_4.png}
    \includegraphics[width=0.3\textwidth]{Styles/pictures/CIFAR/comparison_automobile_5.png}
    \includegraphics[width=0.3\textwidth]{Styles/pictures/CIFAR/comparison_automobile_32.png}
    
    \includegraphics[width=0.3\textwidth]{Styles/pictures/CIFAR/comparison_bird_6.png}
    \includegraphics[width=0.3\textwidth]{Styles/pictures/CIFAR/comparison_bird_13.png}
    \includegraphics[width=0.3\textwidth]{Styles/pictures/CIFAR/comparison_bird_18.png}
    
    \includegraphics[width=0.3\textwidth]{Styles/pictures/CIFAR/comparison_cat_9.png}
    \includegraphics[width=0.3\textwidth]{Styles/pictures/CIFAR/comparison_cat_17.png}
    \includegraphics[width=0.3\textwidth]{Styles/pictures/CIFAR/comparison_cat_21.png}
    
    \includegraphics[width=0.3\textwidth]{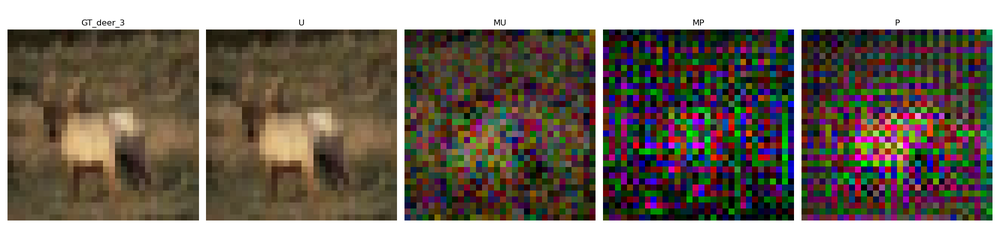}
    \includegraphics[width=0.3\textwidth]{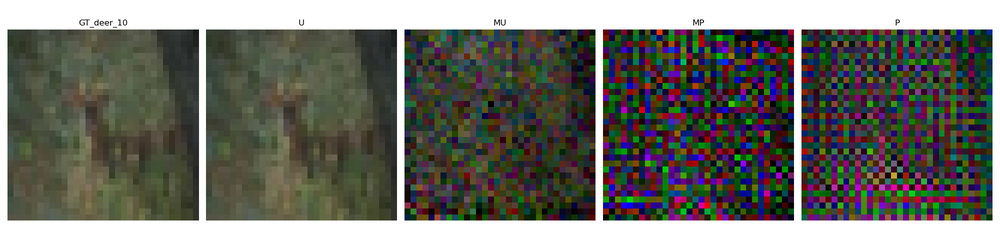}
    \includegraphics[width=0.3\textwidth]{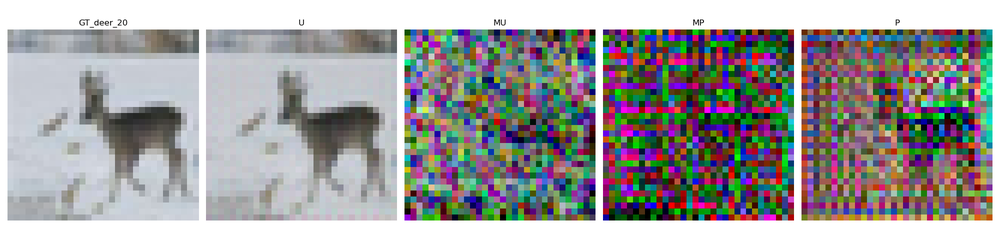}
    
    \includegraphics[width=0.3\textwidth]{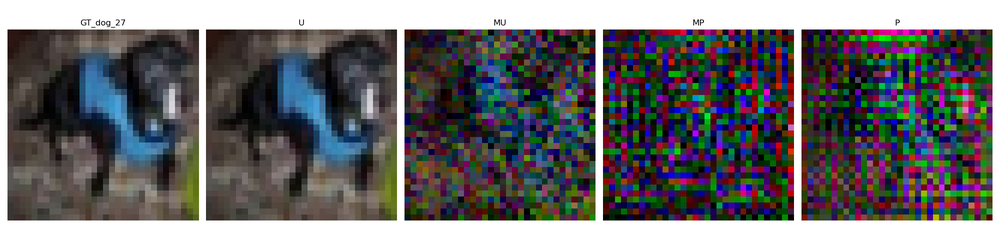}
    \includegraphics[width=0.3\textwidth]{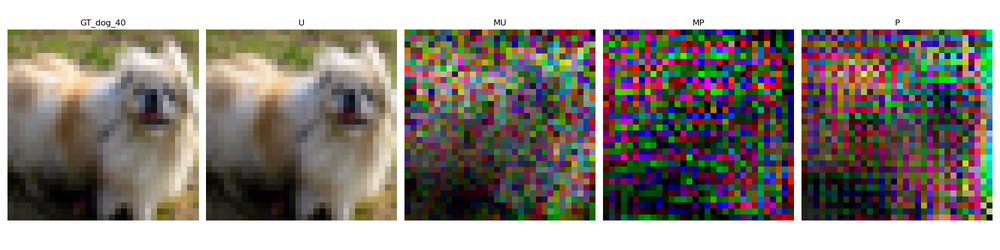}
    \includegraphics[width=0.3\textwidth]{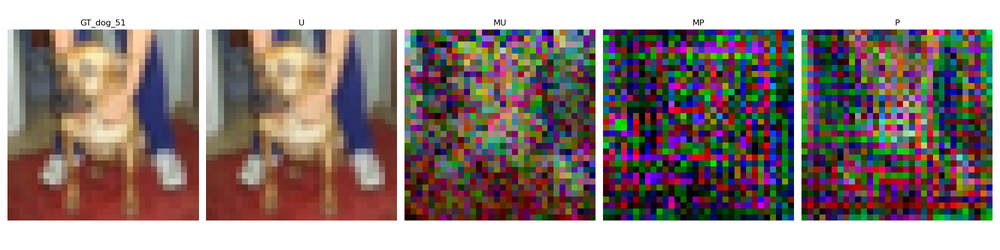}
    
    \includegraphics[width=0.3\textwidth]{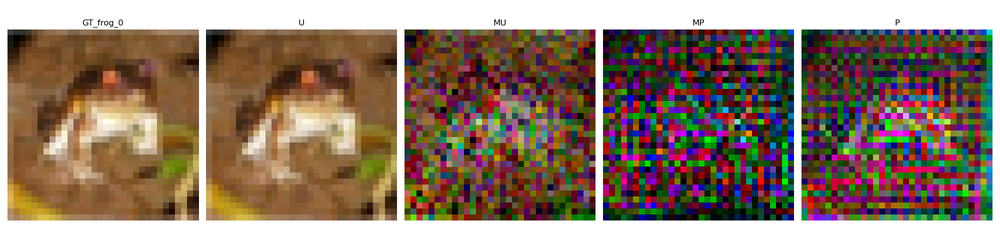}
    \includegraphics[width=0.3\textwidth]{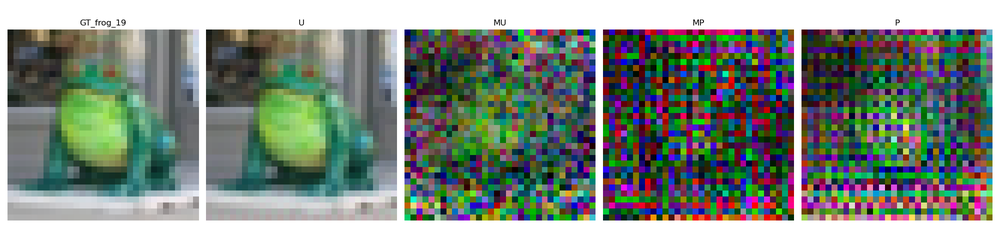}
    \includegraphics[width=0.3\textwidth]{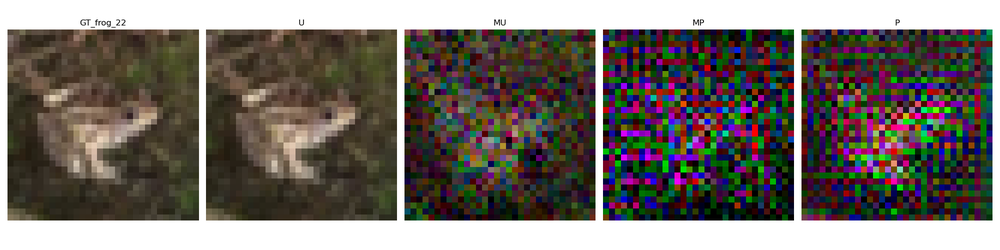}
    
    \includegraphics[width=0.3\textwidth]{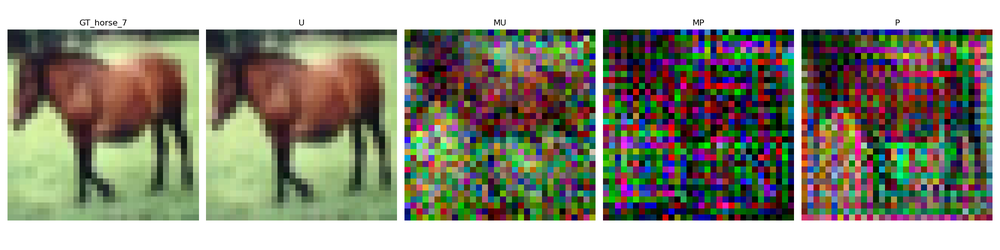}
    \includegraphics[width=0.3\textwidth]{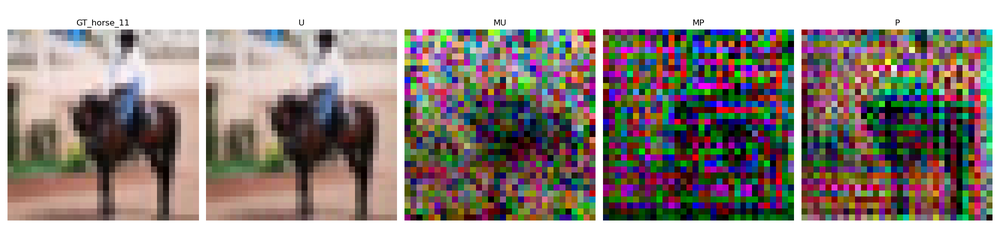}
    \includegraphics[width=0.3\textwidth]{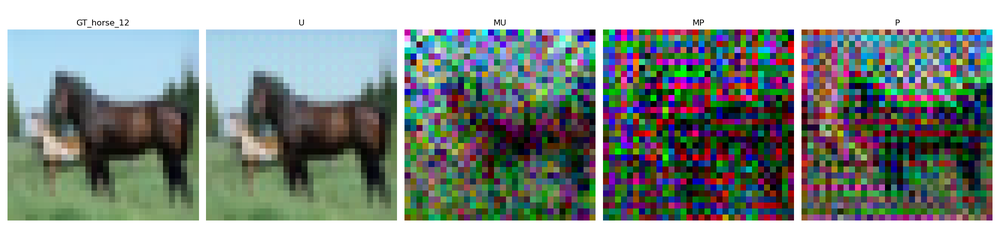}
    
    \includegraphics[width=0.3\textwidth]{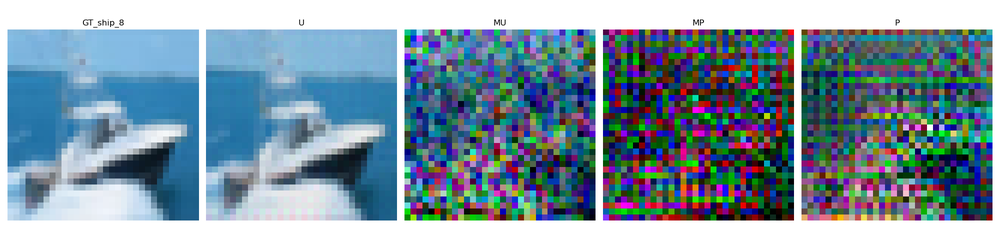}
    \includegraphics[width=0.3\textwidth]{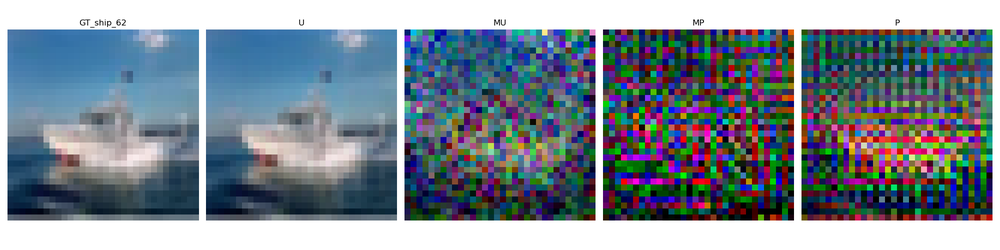}
    \includegraphics[width=0.3\textwidth]{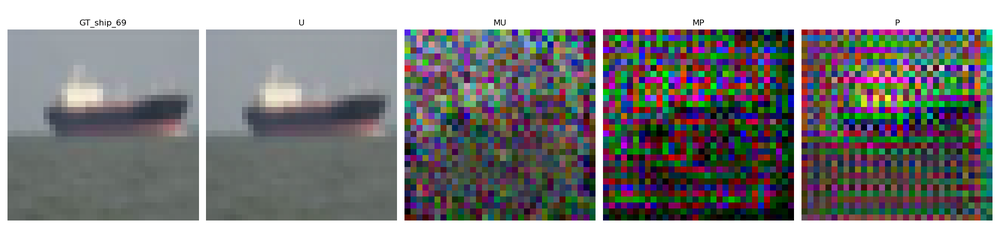}
    
    \includegraphics[width=0.3\textwidth]{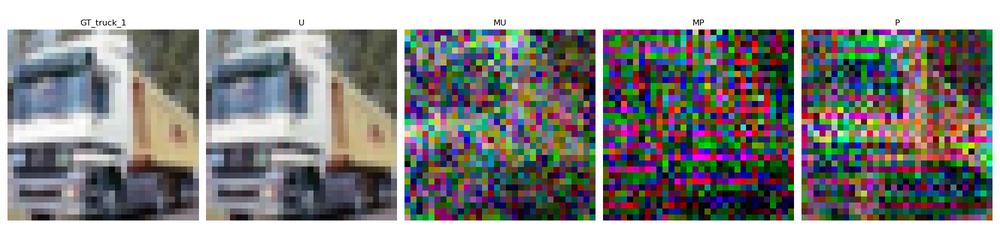}
    \includegraphics[width=0.3\textwidth]{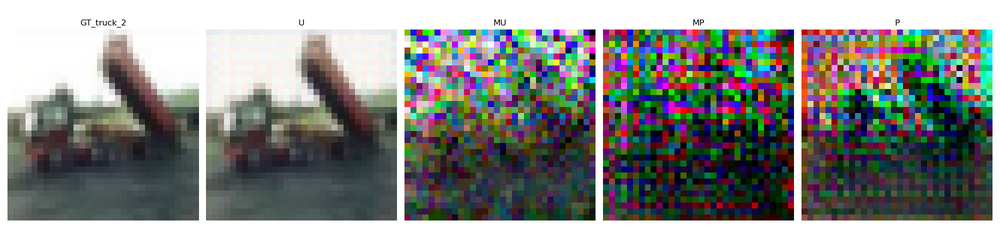}
    \includegraphics[width=0.3\textwidth]{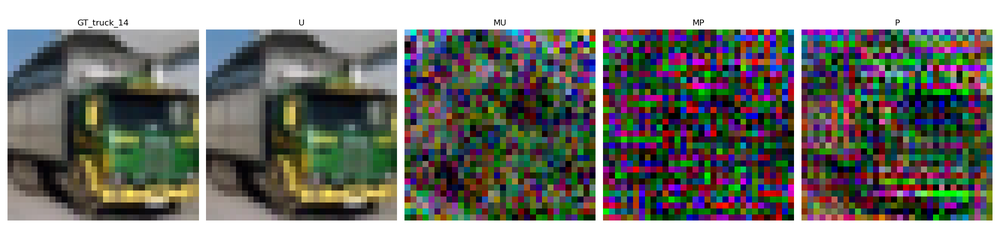}

    \caption{Images from various classes of the \textbf{CIFAR} dataset are displayed, covering the categories Airplane, Automobile, Bird, Cat, Deer, Dog, Frog, Horse, Ship, and Truck (in order from top to bottom). Each row presents three samples from these categories, with different recovery configurations alongside their corresponding ground truth. Similar to Figure \ref{fig:reconstruction_comparisons}, the order for each sample is:\textbf{ GROUND TRUTH, PEEL (U), PEEL (MU), PEEL (P), and PEEL (MP)}.}
    \label{fig:all_cifar_comparisons}
\end{figure}

\begin{figure}[h]
    \centering
    \includegraphics[width=0.3\textwidth]{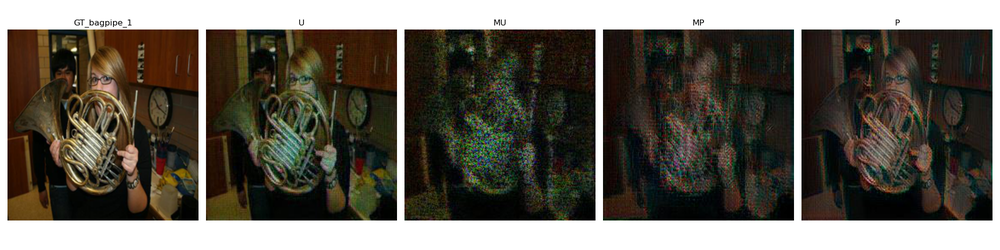}
    \includegraphics[width=0.3\textwidth]{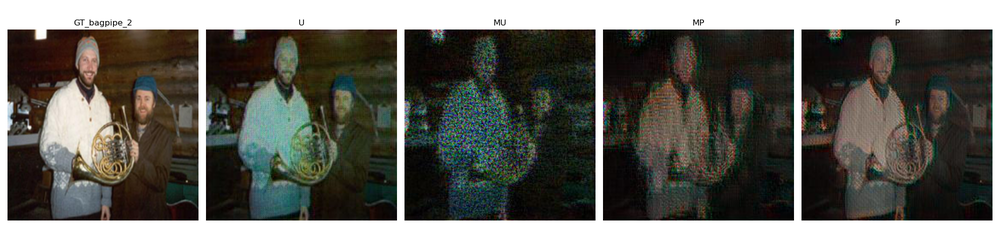}
    \includegraphics[width=0.3\textwidth]{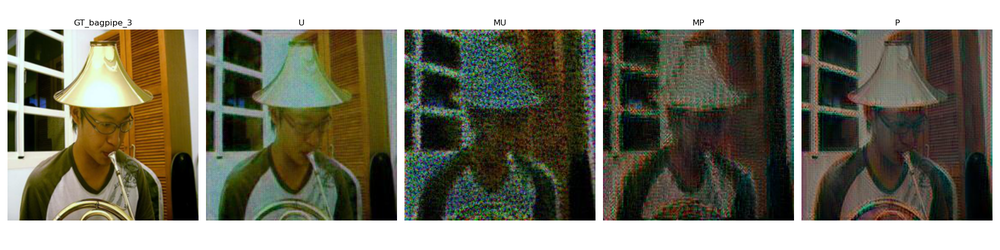}
    
    \includegraphics[width=0.3\textwidth]{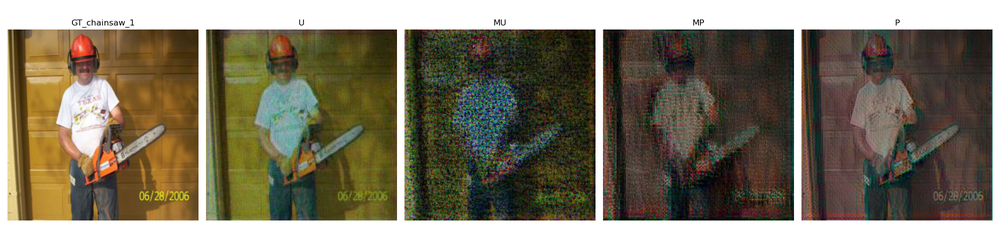}
    \includegraphics[width=0.3\textwidth]{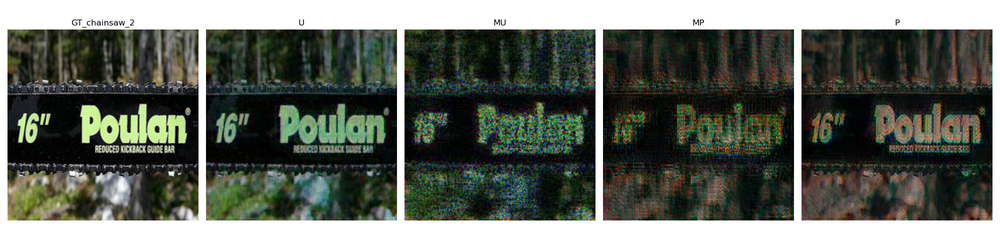}
    \includegraphics[width=0.3\textwidth]{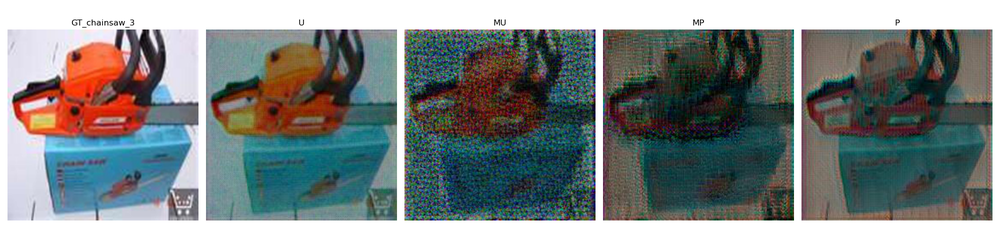}
    
    \includegraphics[width=0.3\textwidth]{Styles/pictures/CIFAR/ImageNet/comparison_church_1.png}
    \includegraphics[width=0.3\textwidth]{Styles/pictures/CIFAR/ImageNet/comparison_church_2.png}
    \includegraphics[width=0.3\textwidth]{Styles/pictures/CIFAR/ImageNet/comparison_church_3.png}
    
    \includegraphics[width=0.3\textwidth]{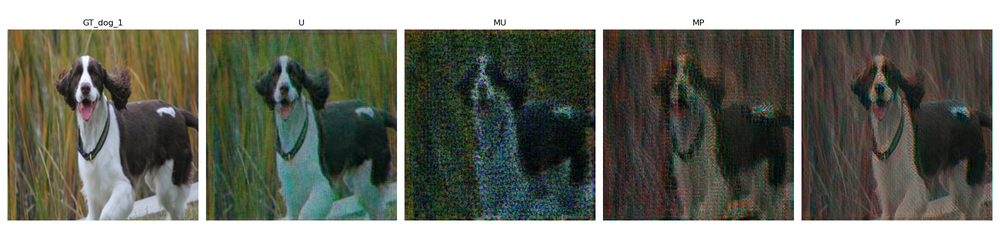}
    \includegraphics[width=0.3\textwidth]{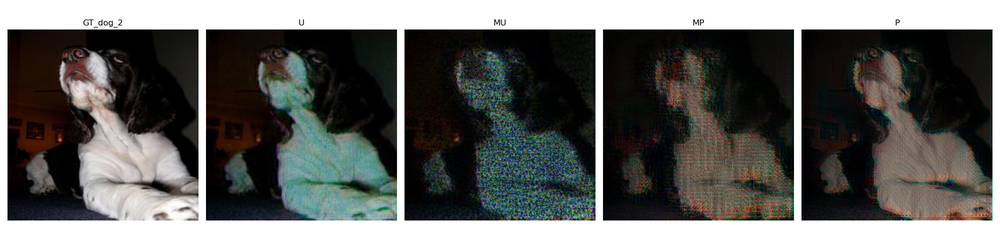}
    \includegraphics[width=0.3\textwidth]{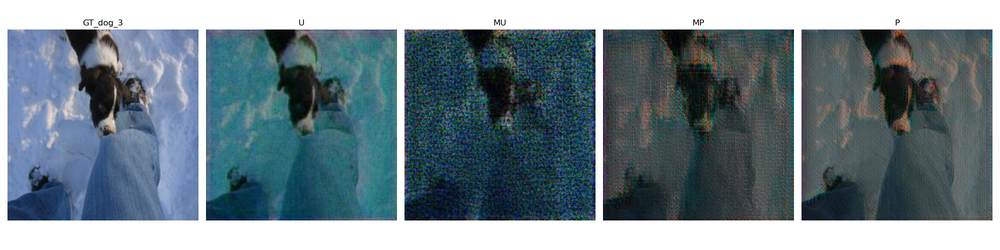}
    
    \includegraphics[width=0.3\textwidth]{Styles/pictures/CIFAR/ImageNet/comparison_fish_1.png}
    \includegraphics[width=0.3\textwidth]{Styles/pictures/CIFAR/ImageNet/comparison_fish_2.png}
    \includegraphics[width=0.3\textwidth]{Styles/pictures/CIFAR/ImageNet/comparison_fish_3.png}
    
    \includegraphics[width=0.3\textwidth]{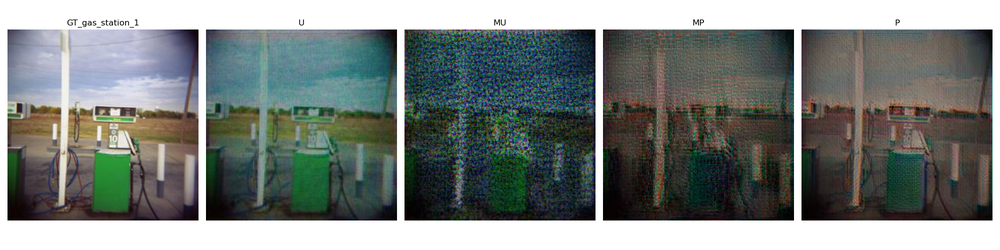}
    \includegraphics[width=0.3\textwidth]{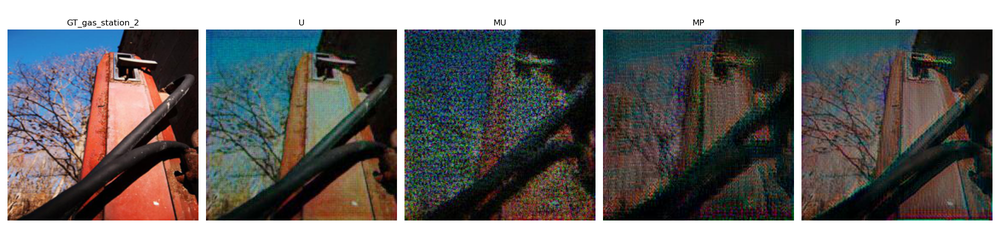}
    \includegraphics[width=0.3\textwidth]{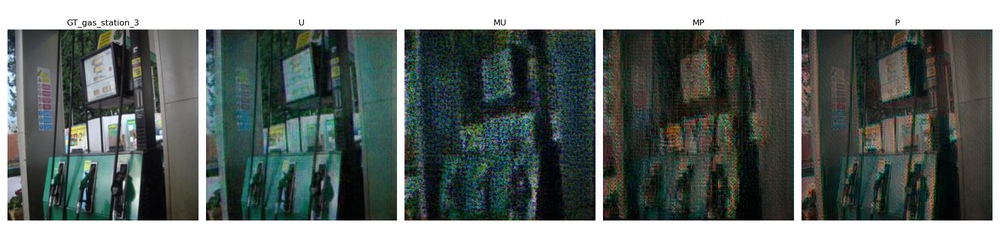}
    
    \includegraphics[width=0.3\textwidth]{Styles/pictures/CIFAR/ImageNet/comparison_golf_ball_1.png}
    \includegraphics[width=0.3\textwidth]{Styles/pictures/CIFAR/ImageNet/comparison_golf_ball_2.png}
    \includegraphics[width=0.3\textwidth]{Styles/pictures/CIFAR/ImageNet/comparison_golf_ball_3.png}
    
    \includegraphics[width=0.3\textwidth]{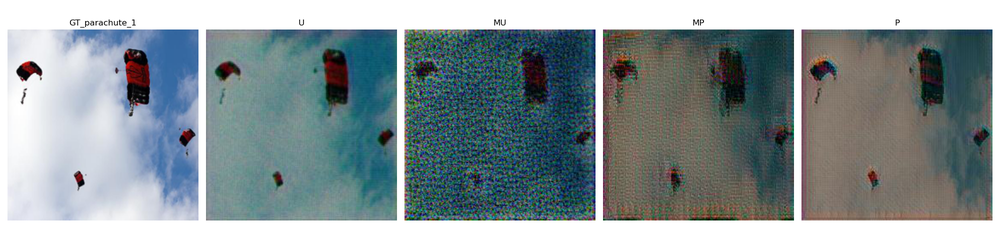}
    \includegraphics[width=0.3\textwidth]{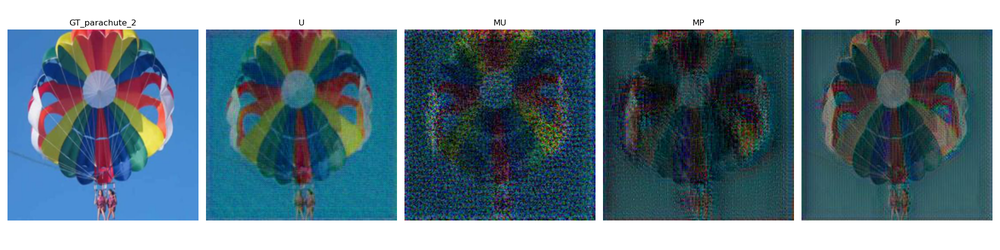}
    \includegraphics[width=0.3\textwidth]{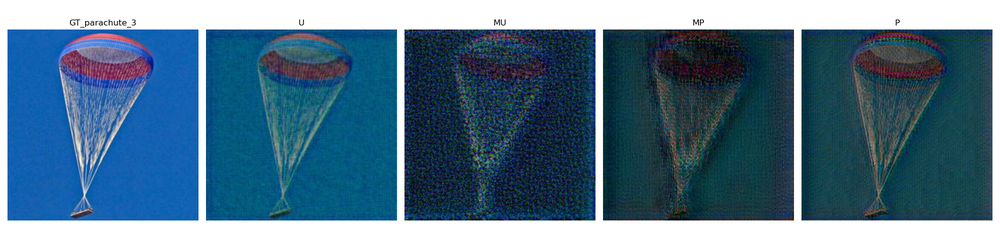}
    
    \includegraphics[width=0.3\textwidth]{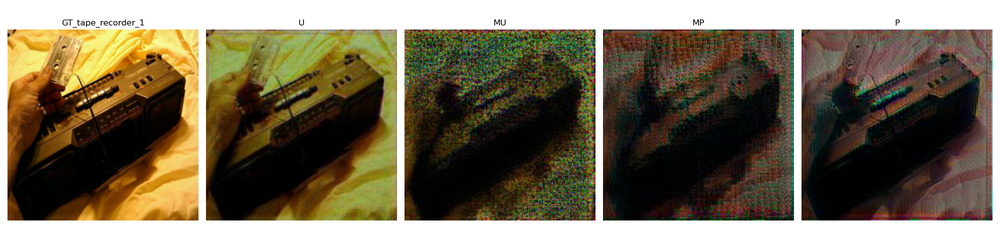}
    \includegraphics[width=0.3\textwidth]{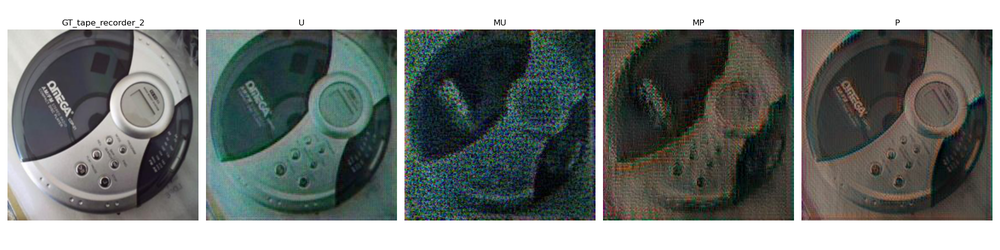}
    \includegraphics[width=0.3\textwidth]{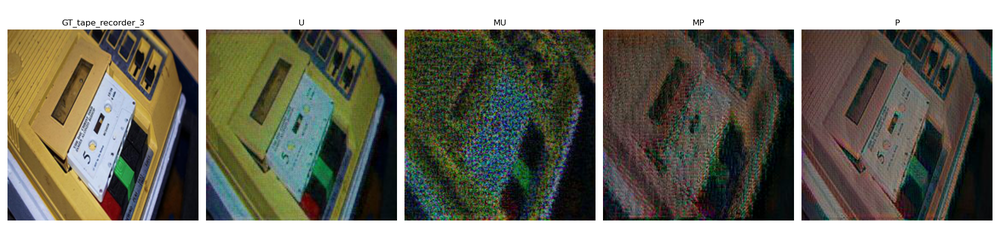}
    
    \includegraphics[width=0.3\textwidth]{Styles/pictures/CIFAR/ImageNet/comparison_truck_1.png}
    \includegraphics[width=0.3\textwidth]{Styles/pictures/CIFAR/ImageNet/comparison_truck_2.png}
    \includegraphics[width=0.3\textwidth]{Styles/pictures/CIFAR/ImageNet/comparison_truck_3.png}

    \caption{Images from various classes of the \textbf{ImageNet} dataset are presented, specifically from the following categories:  Bagpipe, Chainsaw, Church, Dog, Fish, Gas Station, Golf Ball, Parachute, Tape Recorder, and Truck (top to bottom). Each row displays three samples from these classes, with different recovery configurations shown alongside the ground truth. Similar to Figure \ref{fig:reconstruction_comparisons}, the order for each sample is:\textbf{ GROUND TRUTH, PEEL (U), PEEL (MU), PEEL (P), and PEEL (MP)}.}
    \label{fig:all_imagenet_comparisons}
\end{figure}

Figures \ref{fig:all_cifar_comparisons} present additional examples of input reconstructions from the CIFAR dataset, showcasing various classes and configurations. Similarly, Figure \ref{fig:all_imagenet_comparisons}
provides further examples of reconstructions from the ImageNet dataset, illustrating the effectiveness of \textbf{PEEL} across diverse distributions.

\section*{L. IR-152 Model}
\label{sec:IR-152}

The IR-152 model \cite{Wang2022FaceEvoLVe}, a deep convolutional neural network, is specifically designed for face recognition tasks and is widely employed in generative methods for facial reconstruction. This model belongs to the InsightFace ResNet (IR) series, which features architectures with varying depths, such as IR-50, IR-101, and IR-152, where the number indicates the network’s layer count. In alignment with \cite{KEDMI} and \cite{rethink}, we trained the IR-152 model on the CelebA dataset for the purposes of our study.

\section*{M. Additional Discussion of the HbC Setting}
\label{sec:additional HbC}

In the HbC scenario we considered, courts could view attempts by a service provider to secretly reconstruct user data from cached outputs as a breach of trust and legal agreements. However, from a security and cryptography standpoint, the \emph{honest-but-curious (HbC)} adversary model remains a valuable theoretical framework for analyzing potential privacy risks. Indeed, the HbC model has been widely adopted in research on gradient leakage in federated learning \cite{zhu2019deep,malekzadeh2021honest} and other secure computation scenarios \cite{paverd2014modelling}, as it exposes vulnerabilities that may arise even under formally compliant behavior. By assuming an adversary that follows the agreed protocols but seeks to infer as much information as possible, one can design robust systems that defend against both inadvertent leakage and deliberate misuse.

Moreover, unintended disclosures can occur even without malicious intent, due to factors like caching misconfigurations or inadequate access controls. Historical examples include the 2017 ``Cloudbleed'' incident---where a bug in Cloudflare’s code caused private memory contents to leak through HTTP responses \cite{cloudbleed2017}---and Web-Cache Deception Attacks (WCD), in which improperly cached URLs allowed attackers to retrieve private data from high-profile websites \cite{mirheidari2020cached}. Analyzing such scenarios through the HbC lens helps highlight design flaws that could compromise privacy.

\paragraph{Split Learning and MPC Scenarios.}
Beyond a purely theoretical lens, similar threats can arise in \emph{split learning} or \emph{secure multi-party computation (MPC)} contexts. In split learning, for instance, a neural network is split between the client (data owner) and the server (service provider), and intermediate activations are exchanged to complete forward/backward passes. Certain setups allow the server white-box visibility into not only its part of the network but also the client’s architecture and parameters; relevant works include ARES \cite{samikwa2022ares} and CRSFL \cite{wazzeh2024crsfl}, which examine scenarios where the server can indeed access these intermediate representations \cite{wang2024mpc}. Similarly, in MPC-based model inference \cite{trieflinger2023carbyne,goyal2021atlas}, the server may know the global model architecture (and even weights) without seeing the client’s raw data. Under these conditions, an HbC adversary who lawfully receives intermediate outputs can still attempt to invert them to recover sensitive inputs. By considering such risks \emph{a priori}, one can build stricter protocols and better safeguards to mitigate potential leakage.

\section*{How to adapt PEEL to new residual architectures?}
\label{ref:adapt_new_PEEL}

PEEL is designed to invert residual blocks by leveraging their inherent skip connections, making it adaptable to a variety of residual-based models (e.g., ResNet-18, ResNet-50, ResNet-152). To generalize PEEL to other architectures, follow these steps:

\begin{enumerate}
    \item \textbf{Understand Residual Block Structure:} Analyze the target network’s residual blocks, noting variations such as the number of layers, bottleneck layers, and activation functions.

    \item \textbf{Adjust Optimization Problem:} Modify the formulation to account for:
    \begin{itemize}
        \item Additional convolutional layers or bottlenecks.
        \item Properties of alternative activation functions (e.g., Leaky ReLU).
    \end{itemize}

    \item \textbf{Update Constraints:} Revise optimization constraints to reflect the target network’s block structure, including strides and dilations.
\end{enumerate}

\end{document}